\documentclass[10pt,twocolumn,letterpaper]{article}

\usepackage[pagenumbers]{iccv}              

\usepackage{times}
\usepackage{epsfig}
\usepackage{graphicx}
\usepackage{xcolor, soul}
\usepackage{comment}
\usepackage{url}
\usepackage{wrapfig}
\usepackage{lipsum} 
\usepackage{float}

\usepackage{amsmath}
\usepackage{amsthm}
\usepackage{amssymb}
\usepackage{pifont} 
\usepackage{calrsfs}
\usepackage{nccmath}
\usepackage{mathtools, cases}

\usepackage{algorithm}
\usepackage{algorithmic}

\usepackage{booktabs}
\usepackage{multirow}
\usepackage{hhline, colortbl}
\usepackage{subcaption}
\usepackage{ctable}

\definecolor{iccvblue}{rgb}{0.21,0.49,0.74}
\usepackage[pagebackref,breaklinks,colorlinks,allcolors=iccvblue]{hyperref}

\sethlcolor{red}

\def\paperID{5671} 
\def\confName{ICCV}
\def\confYear{2025}
\newcommand{\cmark}{\footnotesize \color{blue} \ding{51}}%
\definecolor{awesomePINK}{rgb}{1.0, 0.13, 0.32}
\definecolor{awesomeGRAY}{rgb}{0.5,0.5,0.5}
\definecolor{awesomeYELLOW}{rgb}{0.99, 0.93, 0.0}
\definecolor{TableYELLOW}{rgb}{0.98, 0.91, 0.71}

\definecolor{Gray}{gray}{0.85}
\definecolor{LightCyan}{rgb}{0.88,1,1}

\def\paperID{5671} 
\def\confName{ICCV}
\def\confYear{2025}

\title{Mixture of Experts Guided by Gaussian Splatters Matters: A new Approach to Weakly-Supervised Video Anomaly Detection}

 \author{\parbox{16cm}{\centering
    {\large Giacomo D'Amicantonio$^{1{\color{red}*}}$, Snehashis Majhi$^{2,3{\color{red}*}}$, Quan Kong$^4$, Lorenzo Garattoni$^4$,  Gianpiero Francesca$^4$, François Brémond$^{2,3}$, Egor Bondarev$^1$}\\
    {\small
    $^1$ Eindhoven University of Technology \quad
    $^2$ INRIA \quad
    $^3$ Côte d'Azur University \quad
    $^4$ Woven by Toyota\quad
    $^5$ Toyota Motor Europe
    }}\\
    \vspace{-0.3cm}
    \small{{\color{red}* Joint first authors.}} \quad \small{{{Code: \url{https://github.com/snehashismajhi/GS-MoE}}}}
}

\begin{document}
\maketitle
\begin{abstract}

Video Anomaly Detection (VAD) is a challenging task due to the variability of anomalous events and the limited availability of labeled data. Under the Weakly-Supervised VAD (WSVAD) paradigm, only video-level labels are provided during training, while predictions are made at the frame level. Although state-of-the-art models perform well on simple anomalies (e.g., explosions), they struggle with complex real-world events (e.g., shoplifting). This difficulty stems from two key issues: (1) the inability of current models to address the diversity of anomaly types, as they process all categories with a shared model, overlooking category-specific features; and (2) the weak supervision signal, which lacks precise temporal information, limiting the ability to capture nuanced anomalous patterns blended with normal events. To address these challenges, we propose \textbf{Gaussian Splatting-guided Mixture of Experts (GS-MoE)}, a novel framework that employs a set of expert models, each specialized in capturing specific anomaly types. These experts are guided by a temporal Gaussian splatting loss, enabling the model to leverage temporal consistency and enhance weak supervision. The Gaussian splatting approach encourages a more precise and comprehensive representation of anomalies by focusing on temporal segments most likely to contain abnormal events. The predictions from these specialized experts are integrated through a mixture-of-experts mechanism to model complex relationships across diverse anomaly patterns. Our approach achieves state-of-the-art performance, with a 91.58\% AUC on the UCF-Crime dataset, and demonstrates superior results on XD-Violence and MSAD datasets. By leveraging category-specific expertise and temporal guidance, GS-MoE sets a new benchmark for VAD under weak supervision.
\end{abstract}    
\section{Introduction}
\label{sec:intro}

Video Anomaly Detection (VAD) in surveillance videos is one of the most challenging tasks in the field of Computer Vision. With the increasing capabilities of deep-learning models, there have been various approaches to tackle this task. The main focus of recent research in the field of VAD has been to model spatio-temporal dependencies in videos, obtaining meaningful representations of the motion of relevant agents in the scene. In this sense, the transformer architecture has proved to be very effective, forming the backbone of multiple works. While the current state-of-the-art models have achieved reasonable results on publicly available datasets, they still fail to capture subtle anomalies and to detect the temporal window in which they happen. 

We identify one of the main reasons for these issues in the formulation of the WSVAD task~\cite{sultani2018real, wu2022weakly}. Multi-instance learning (MIL) strikes a balance between fully supervised methods, which exhibit good performance but require costly data annotation, and unsupervised methods, which do not require manual annotations but generally result in worse performance. The core idea of MIL is to create bags containing positive and negative data samples (\textit{i.e.}, normal and abnormal videos), labeled only at the video-level. During training, the model assigns a score between 0 and 1 to each snippet, with 0 indicating a normal snippet and 1 indicating an abnormal snippet. The highest-scoring samples in the normal bag are guided towards 0, allowing the model to learn most normal scenarios correctly. On the other hand, the highest-scoring negative samples are pushed towards 1. This leads the model to be supervised, and therefore learn, few and specific instances of anomalous events, ignoring useful information contained in neighboring snippets. Over time, this approach has proved to be powerful but insufficient to train a model to correctly capture the secondary and specific attributes of different anomalous classes. In recent works~\cite{yu2020cloze, yan2023towards, georgescu2021anomaly}, different auxiliary objectives are identified as priors for the VAD task to optimize the training process. \\
\indent
\begin{figure}[t]
\centering
\includegraphics[width=\linewidth]{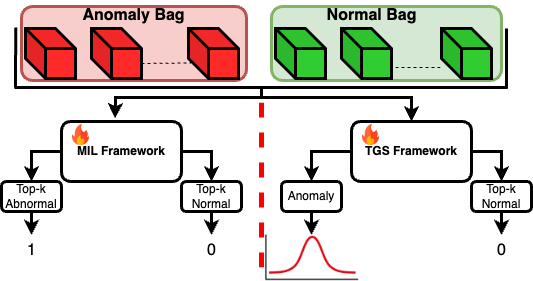}
\caption{While SoTA methods address the task of WSVAD via the most normal and abnormal snippets in a video, the approach proposed in this paper focuses on learning a more complete representation of anomalous events via Gaussian kernels.} 
\label{fig:task}

\end{figure}
To address this issue, we propose to model the anomalies in a video as Gaussian distributions, rendering multiple Gaussian kernels in correspondence with peaks detected along the temporal dimension of the scores estimated for abnormal videos. This technique, called Temporal Gaussian Splatting (TGS), creates a more complete representation of an anomalous event over time, including snippets of the anomaly with lower abnormal scores in the training objective. A side-by-side comparison of the MIL task and the TGS task is shown in Figure~\ref{fig:task}.
The Gaussian kernels are extracted from the abnormal scores produced by the model. \\
An additional challenge is related to the intrinsic differences between abnormal classes. Under the MIL paradigm, the models are trained to learn the difference between normal and abnormal videos, while the specific differences between anomalous classes are overlooked. As a result, these methods mainly focus on coarse-level representations of anomalies that allow us to distinguish between normal and abnormal events, but ignore the fine-grained category-specific cues. Therefore, the more salient anomalies (\textit{i.e.}, such as an explosion) are likely to be easily detected, while subtle anomalies (\textit{i.e.}, shoplifting) are more likely to be confused with normal events. This constitutes a major limitation of most recent methods based on WSVAD. We address this issue via a Mixture-of-Expert (MoE) architecture, in which each expert is trained to model a single anomaly class, enhancing the specific attributes of each anomaly class that are often overlooked. To further leverage the correlations and differences between anomalies, a gate model mediates between the predictions of each expert and the more coarse-level anomalous features to learn potential interactions between anomalies. \\
The contributions of this paper are complementary: learning specific representations of anomalous classes allows for more accurate Gaussian kernels, and the Gaussian splatting enables the experts to learn from more subtle anomalous events that would be overlooked otherwise. To summarize, this paper presents:
\begin{itemize}
    \item A novel formulation of the WSVAD task based on Gaussian kernels extracted from the estimated abnormal scores to generate a more expressive and complete representation of anomalous events. Splatting the kernels along the temporal dimension allows the model to learn more precise temporal dependencies between snippets and highlight more subtle anomalies;
    \item A Mixture-of-Expert (MoE) architecture that focuses on individual anomaly types via dedicated class-expert models, allowing a gate model to leverage similarities and diversities between them;
    \item The impact of the proposed contributions is measured via an extensive set of experiments on the challenging UCF-Crime~\cite{cvpr18}, XD-Violence~\cite{eccv2020} and MSAD~\cite{msad2024} datasets, showing notable improvements in performance w.r.t. previous SoTA methods. 
\end{itemize} 

\section{Related Work}
\paragraph{Weakly-Supervised VAD: }
In the WSVAD task, anomalous events encompass various classes, each exhibiting distinct characteristics across the spatial and temporal dimensions. 
The task of WSVAD was introduced in a seminal work by~\cite{sultani2018real}. In the following years, there have been multiple different approaches that addressed the trade-off between the ease of data collection and the performance exhibited by models trained in this task. The limitation of weak labels was addressed by~\cite{zhong2019graph} using a graph convolutional network to correct noisy labels and supervise traditional anomaly classifiers. Further,~\cite{tian2021weakly} proposed to learn a function of the magnitude of features to improve the classification of normal snippets and, therefore, the detection of abnormal events. The model is based on attention modules and pyramidal convolutions. The idea of improving the quality of weak labels was also explored by~\cite{li2022self}, which designed a transformer-based method trained to predict abnormal scores both at the snippet and video levels. The video-level predictions are then used to improve the performance of the model at the snippet level. More recently,~\cite{zhang2023exploiting} designed a multi-head classification model that leveraged uncertainty and completeness to produce and refine its own pseudo-labels.~\cite{majhi2024oe} proposed a two-stage transformer-based model that generates anomaly-aware position embeddings and then models the short and long-range relationships of anomalous events. Inspired by point-supervision~\cite{bearman2016s},~\cite{zhang2024glancevad} introduced Glance annotations. These annotations enhance the common weak labels by localizing a single frame in which an anomalous event is happening. While reporting very good performance, these annotations require an additional manual-labelling procedure. Recently,~\cite{Majhi_2025_CVPR} proposed a method to include additional data modalities in the anomaly detection process.\\
Under the MIL paradigm, these variations complicate the model’s ability to effectively differentiate between them. By focusing on the $\text{top}_k$ most abnormal snippets of a video, the model is guided towards specific and evident anomalous events, without properly considering the sequence of actions that lead to them and follow them. In fact, some anomalies occur within short time windows, while others unfold over longer periods; moreover in both cases, the MIL paradigm selects the same amount of abnormal snippets. 

\paragraph{Mixture of Experts:}
This architecture has been introduced by~\cite{eigen2013learning} and has since been improved and employed for diverse tasks, from image classification to action recognition~\cite{jain2024mixture}. The original MoE design proposed a series of small experts and a separate gate network, all receiving the same input data. Each expert predicts an output, while the gate network assigns a score of importance to them. Since then, this architecture has been improved upon by various works. A common idea across domains is to let a routing network select which portions of the input data, or input tokens, to pass to each expert~\citep{riquelme2021scaling, mustafa2022multimodal, fedus2022switch, lepikhin2020gshard}. A recent work by
\cite{puigcerver2024sparsesoftmixturesexperts} proposed to weight the input tokens in a different way for each expert.

\paragraph{Gaussian Splatting:}It has received a lot of attention in recent years, proving to be very efficient in fields like 3D scene reconstruction~\cite{kerbl20233d, kopanas2021point}. The main idea of Gaussian Splatting is to represent each 3-dimensional point in a scene as a multivariate normal distribution, which allows to render the scene as the sum of the contributions of all the 3-dimensional areas.
Gaussian splatting has since been extended to incorporate the temporal dimension in multiple domains, for example, dynamic scene rendering~\cite{li2024spacetime, 4DGS} and medical imaging~\cite{zhang2024togsgaussiansplattingtemporal}. 

Our approach utilizes Mixture-of-Experts (MoE) by assigning each expert to a specific anomaly class, enabling fine-grained, category-specific learning often missed in traditional designs. A gate model bridges class-specific experts and coarse features, leveraging anomaly correlations while ensuring balanced utilization. We extend Gaussian Splatting into the temporal domain with Temporal Gaussian Splatting (TGS), capturing nuanced dependencies and integrating subtle, low-scoring snippets into training. Anchored to temporal peaks, TGS mitigates noise, enhances weak supervision, and preserves sharp transitions, avoiding over-smoothing while delivering precise anomaly detection.

\section{Methodology}
Our novel Gaussian Splatter-guided Mixture-of-Experts (GS-MoE) framework aims to accurately detect complex anomalies using weakly-labeled training videos. GS-MoE leverages two key techniques: \textbf{(I) Temporal Gaussian Splatting loss}, to ensure superior separability between normal and anomalous instances under weak-supervision; \textbf{(II) Mixture-of-Experts (MoE) architecture}, that learns class-specific representations and detects complex anomalies with high confidence. 
\subsection{Temporal Gaussian Splatting (TGS)}
\begin{figure}[t]
\begin{center}
\includegraphics[width=.5\textwidth]{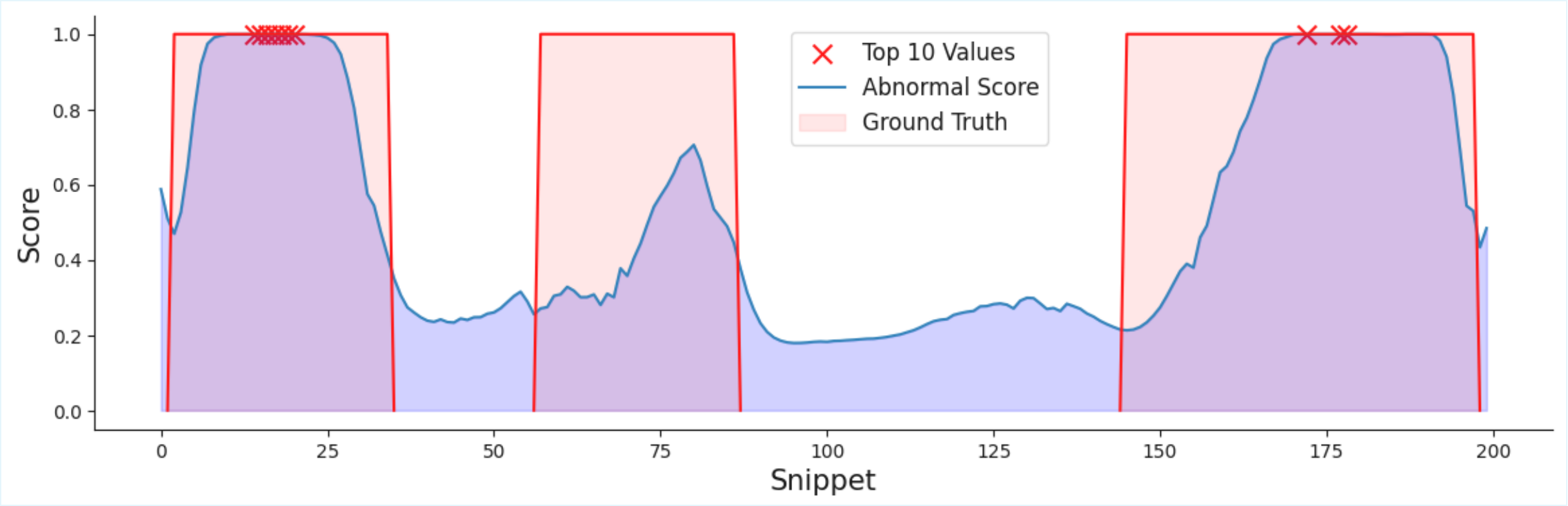}
\caption{The abnormal scores obtained from the backbone model on a training video at the end of training. The $top_k$ snippets used in the MIL paradigm lead the model to focus on the first and last of the three anomalous events present in the video, overlooking the second anomaly. However, the second anomaly, while not scoring as high as the others, is still detected. }
\label{fig:abn_score}
\end{center}
\end{figure}

Our Temporal Gaussian Splatting (TGS) technique provides a novel formulation of the MIL optimization paradigm by leveraging Gaussian kernels. The core idea of TGS is to reduce the over-dependency on the most abnormal snippets that is often the result of the classical MIL.
An example of such over-dependency is shown in Figure~\ref{fig:abn_score}. The $\text{top}_k$ abnormal scores are the ones that would normally be used in the loss function in the MIL paradigm:
\begin{equation}
\text{top}_k(S) = \{score_{1}, score_{2}, \dots, score_{k}\} \quad
\label{topk}
\end{equation}

such that 

\begin{align}
    \setlength{\jot}{0pt} 
    L_{\text{top}_k} =\ & - \underbrace{\frac{1}{N^+} \sum_{i=1}^{N^+} \frac{1}{k} \sum_{j \in \text{top}_k(\mathbf{S^+})} \log \sigma(\text{score}_{ij})}_{L_{\text{top}_k-\text{abn}}} + \nonumber \\
    & \underbrace{- \frac{1}{N^-} \sum_{i=1}^{N^-} \frac{1}{k} \sum_{j \in \text{top}_k(\mathbf{S^-})} \log \left(1 - \sigma(\text{score}_{ij})\right)}_{L_{\text{top}_k-\text{norm}}}
    \label{loss}
\end{align}
where $S \in {S^+, S^-}$ and  \(\text{score}_{1} \geq \text{score}_{2} \geq \dots \geq \text{score}_{k}\).  Here, \(S^-\) and  \(S^+\) denote the set of scores derived from an abnormal and normal video, and \(\text{score}_{i}, \, i \in \{1, \dots, k\}\) represents the score of the \(i\)-th highest-ranked snippet.
Similarly, $N^+$ and $N^-$ are the number of videos in the abnormal and normal classes and $\sigma$ is the sigmoid function.

\begin{figure}[t]
\begin{center}
\includegraphics[width=.5\textwidth]{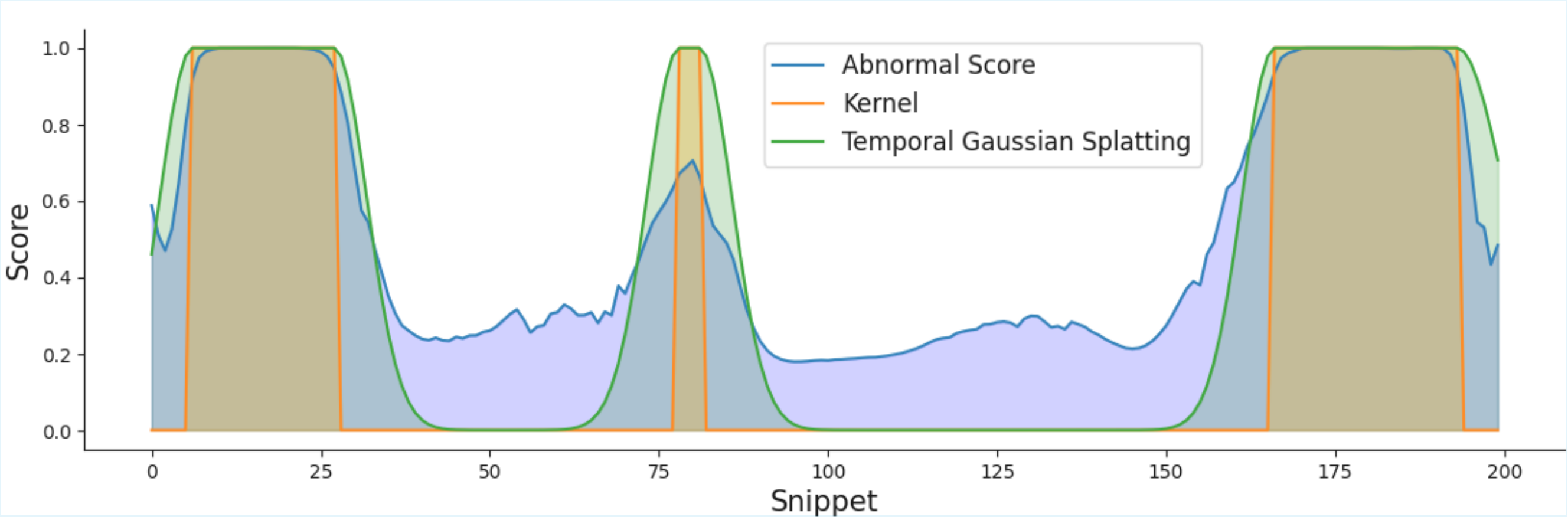}

\caption{The Gaussian kernels extracted from the abnormal scores shown in Figure~\ref{fig:abn_score} are splatted across the width of the detected peaks. This allows the model to learn a more complete representation of the anomalous events in the video.}
\label{fig:splat1}
\end{center}
\end{figure}
\begin{figure*}[t]
\begin{center}
\includegraphics[width=1.\textwidth]{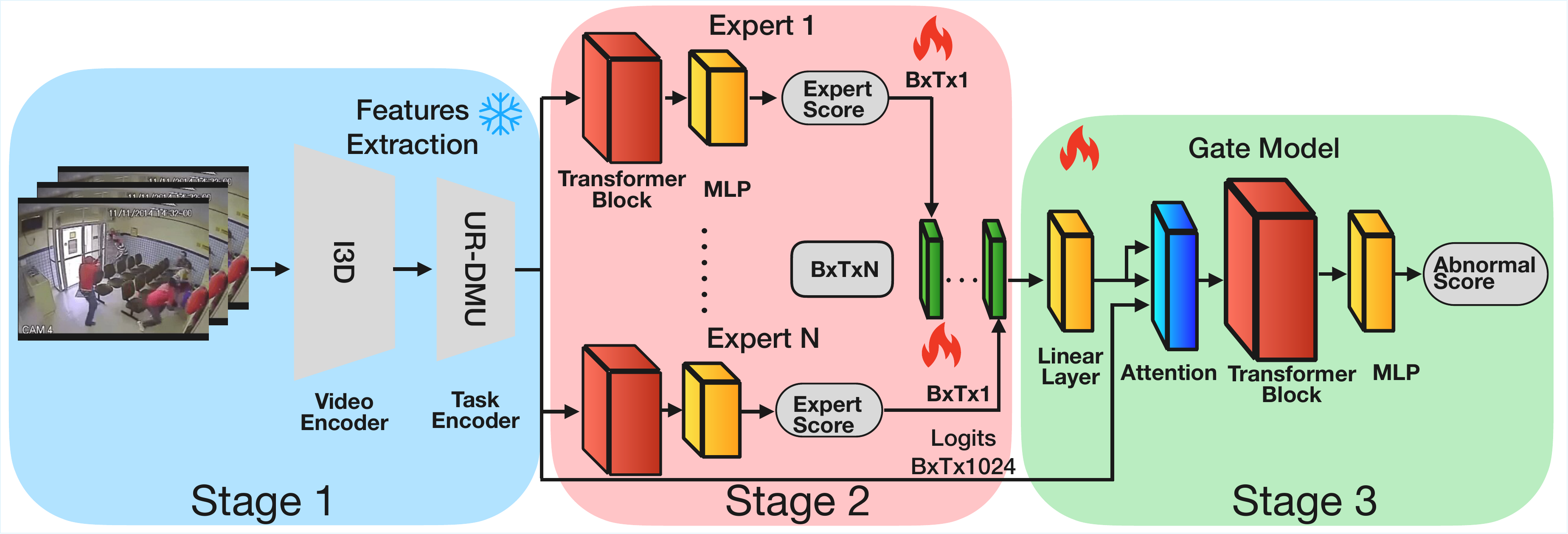}
\caption{\textbf{Overview of the GS-MoE architecture:} First, in the feature extraction stage, the video encoder extracts snippet-level features from the video, and the task encoder refines them in the anomaly-detection latent space. In the second stage, each class-expert is trained only on refined features belonging to its assigned class and to the normal class. In the final stage, the gate model collects the scores assigned by each expert and compares them with the refined features of the task encoder, producing the final abnormal score.}
\label{fig:architecture_gate}
\end{center}
\end{figure*}

At the end of the training, the task encoder is able to detect two out of three anomalies contained in the video as in Figure~\ref{fig:abn_score}, assigning a very high abnormal score to most snippets in the first and third anomalies time window. The model is not as confident about the snippets belonging to the second anomaly, due to the fact that during training, it has never been supervised specifically on them, but it assigns them an anomalous score higher than the normal snippets of the video. Additionally, the snippets between anomalies are still considered partially anomalous. 
We conjecture that it is possible to leverage those situations to generate pseudo-labels that allow a model to be trained on more information, while remaining in the data-annotation boundaries of the WSVAD paradigm. Following~\cite{zhang2024glancevad}, we propose a technique called Temporal Gaussian Splatting (TGS) that leverages the peaks in the abnormal scores predicted by the model to represent precisely the temporal windows in which an anomalous event is happening.

\paragraph{Peak detection:} Gaussian kernels are extracted in correspondence with local maxima, called "peaks", in the temporal axis of the abnormal scores predicted by a model. The peaks are detected by thresholding the local maxima, selecting only the ones that are above a minimum prominence threshold over the previous two scores and the subsequent two. The width $W_i$ of each peak $P_i$ is determined by $minimum(v_1,v_2)$, where $v_1$ is the number of preceding snippets with monotonically increasing scores, and $v_2$ is the number of following snippets with monotonically decreasing scores.

The set of peaks $P$, detected for a given video, contains the position of the snippet with the highest abnormal score for each peak in the video. This may lead to the detection of spurious peaks, meaning peaks in the abnormal scores of a video that do not belong to an anomalous event. To mitigate this, the model can be trained for a few iterations with the $L_{topk-norm}$ component of the standard MIL training objective. 

This allows us to identify subtle anomalies that are usually not included in the $\text{top}_k$ snippets described in Equation \ref{loss}. The kernels obtained from the detected peaks are then rendered over the length of anomalous videos to obtain a more accurate representation of the anomalies along the temporal dimension.

Gaussian kernels $G_i$ are then initialized with a unitary value for the snippets corresponding to each peak $P_i$ detected in the abnormal scores of the video. To further represent the duration of the anomaly, the kernel values corresponding to snippets that are within the width $W_i$ of the respective peak are also set to 1 if their abnormal score is higher than the difference between the peak score and the standard deviation of the normal distribution centered in the peak:
\begin{equation}
   G_{i, t} =
    \begin{cases}
        1, & \textit{if t} = P_i,  \\
        1, & \textit{if $s_t$} \geq s_{P_i} - \sigma_i \land t \in W_i \\
        0, & \textit{otherwise}
    \end{cases}, \forall t \in [1, T]
    \label{kernel}
\end{equation}
where $s_t$ is the abnormal score assigned to snippet $t$ and $\sigma_i$ is the standard deviation of the normal distribution centered in peak $i$. This allows to treat each anomaly separately, which is beneficial for the WSVAD task due to the fact that different anomalies have different characteristics along the temporal dimension. Computing the Gaussian kernels in this way represents an improvement upon the $\text{top}_k$ formulation, allowing the model to learn from the entirety of an anomalous event instead of its most abnormal snippets. Each kernel is splatted via:
\begin{equation}
    f_i(t) = G_{i,t} \cdot \exp(-\frac{\lVert t-P_i\rVert ^2}{2\sigma_i^2}),  \forall t \in [1, T]
    \label{gaussian_splatting}
\end{equation}

where $T$ is the length of the video and $\sigma_i$ is the standard deviation of the scores around the peak centered in $P_i$ within the width $W_i$. Finally, the pseudo-labels $\hat{y}$ are generated by rendering each of the $K$ extracted kernels over the length of the video:
\begin{equation}
    \hat{y} = \lVert(\sum_{i=1}^kf_i(t))\rVert
    \label{pseudo}
\end{equation}
An example of such pseudo-label (Temporal Gaussian Splatting) is shown in Figure~\ref{fig:splat1}. The generated pseudo-labels contain a target abnormal score between $0$ and $1$ for each snippet in the video, allowing the model to learn the severity of each abnormal snippet. This represents a relevant improvement over the standard MIL training objective, where only the $\text{top}_k$ snippets are pushed towards $1$ in the training objective, as in Equation~\ref{loss}. Instead, the TGS loss function used to train the experts and the MoE is formulated as:
\begin{equation}
    L_{TGS} = L_{topk-norm} + BCE(y, \hat{y})
\label{tgs}
\end{equation}

\subsection{Mixture of Experts (MoE)}
Our Mixture-of-Experts (MoE) architecture, illustrated in Figure~\ref{fig:architecture_gate}, combines three stages. The first is task-agnostic and task-aware feature extraction, the second has class-specific expertise, and the third provides a novel gate mechanism. This multi-stage framework directly addresses the challenges of weak supervision and anomaly diversity, enabling precise detection of complex anomalous patterns through enriched representations and specialized models.
\paragraph{\underline{Stage 1:} Enhanced Temporal-Spatial Feature Extraction.} The synergy of I3D and UR-DMU forms the backbone of feature extraction. The widely-used I3D model provides task-agnostic general features, capturing basic video dynamics. However, these features lack the specificity needed for detecting intricate spatial and temporal anomalies. To address this, UR-DMU~\cite{zhou2023dual} acts as a task-aware feature extractor tailored to anomaly detection. Trained initially with the standard MIL loss~\cite{AAAI23URDMU} and fine-tuned using our novel Temporal Gaussian Splatting (TGS) loss (Equation~\ref{tgs}), UR-DMU extracts highly informative features by leveraging temporal consistency. These richer features filter motion dynamics and fine-grained temporal patterns crucial for distinguishing between normal and abnormal events. While UR-DMU can coarsely differentiate normal and abnormal events, it cannot fully address the complexities of different anomaly types. 

\paragraph{\underline{Stage 2:} Class-Specific Anomaly Detection with Expert Models.}To overcome the limitations of coarse anomaly detection, our framework incorporates multiple expert models optimized with our TGS loss (Equation~\ref{tgs}), each dedicated to identifying specific types of anomalies. This design introduces a crucial level of specialization, allowing the framework to capture the unique attributes of individual anomaly classes. Each expert model is composed of a transformer block with four self-attention heads and a MLP with GELU activation~\cite{gelu}, which maps the extracted features to an anomaly score for its respective class. These expert models leverage the enriched UR-DMU features, expanding the boundaries of the latent anomaly space. 
These fine-grained, specialized experts enable the model to detect subtle or complex anomalies that may blend seamlessly with normal events, a feat that generic models fail to achieve.

\paragraph{\underline{Stage 3:} Collaborative Integration with the Gate Model.} In the final stage of the framework, the scores generated by the expert models are passed to the gate model, which acts as collaborative integration mechanism. This step ensures that the individual strengths of the experts are harnessed to create a unified representation capable of robust anomaly detection. The gate model comprises of three components: \textbf{(a) Score Refinement:} The expert scores are concatenated and projected into a higher-dimensional space, enriching the representation of class-specific anomaly logits. This projection enables the gate model to handle the intricate variations across anomaly classes effectively. \textbf{(b) Bi-Directional Cross-Attention Module:} To bridge the gap between fine-grained class-specific logits and coarse abnormal logits from the task encoder, the gate model incorporates a bi-directional cross-attention mechanism. This module learns the correlations and contrasts between the expert predictions and the coarse anomaly-aware features, allowing the gate model to leverage detailed, class-specific insights as well as the more general anomaly-aware features. \textbf{(c) Final Prediction:} The refined and integrated features are processed through a transformer block followed by a four-layer MLP, similar to the architecture of the experts, to produce the final anomaly scores. This step ensures that the latent space representation is expressive and well-suited for capturing diverse anomaly patterns.
\begin{table}[ht]
\centering
\setlength\tabcolsep{1.pt} 
\begin{tabular}{ll|cc|cc}
    \toprule
    \textbf{Model} & \textbf{Encoder} & \multicolumn{2}{c|}{\textbf{UCF-Crime}} & \multicolumn{2}{c}{\textbf{XD-Violence}}\\
     & & \textbf{AUC} & \textbf{AUC\textsubscript{A}}  & \textbf{AP} & \textbf{AP\textsubscript{A}} \\
    \midrule
        \rowcolor{lightgray}
        \multicolumn{6}{c}{\textit{SoTA Methods With Multi-modal Features}} \\
        M.A.~\cite{bmvc19}&  C3D &  79.10 & 62.18  &  - & - \\
        HL-Net~\cite{eccv2020} &    I3D&      82.44 &      -&   - &    -  \\
        HSN~\cite{HSN}  &  I3D&  85.45 &  -  & -   & -   \\
        MACIL-SD~\cite{yu2022modality} & I3D+audio & - & - & 83.40 & - \\
        UR-DMU~\cite{AAAI23URDMU} & I3D+audio & - & - & 81.77 & - \\
        TPWNG~\cite{CVPR24TPWNG} & CLIP & 87.79 & -& 83.68& - \\
        VadCLIP~\cite{AAAI2024_vadclip}& CLIP& 88.02 & 70.23 &  84.15& -  \\
        \midrule
        \rowcolor{lightgray}
        \multicolumn{6}{c}{\textit{SoTA Methods With RGB only} Features}\\
        \multirow{2}{*}{ MIL~\cite{cvpr18}}&  C3D & 75.41 & 54.25  &  75.68 & 78.61   \\
        &  I3D &  77.42 &   - & -    &   - \\
        TCN~\cite{icip2019}&  C3D &  78.66 & -    &  - & - \\
        GCN~\cite{cvpr19}&  TSN &    82.12  & 59.02   &  78.64& - \\
        MIST ~\cite{cvpr21} &   I3D&    82.30& -    &  - & -  \\
        Dance-SA~\cite{iccv21_Dance}&  TRN &     85.00  & -  &  - & -\\
        RTFM~\cite{iccv21} &   I3D &      84.30 &   62.96 &  77.81 &   78.57  \\
        CLAV~\cite{CVPR23LAA}&  I3D&  86.10& -   &  - & -\\
        UR-DMU~\cite{AAAI23URDMU}&  I3D& 86.97& 70.81  & 81.66 &  83.94\\
        SSRL~\cite{ECCV2022Scale}&  I3D& 87.43&  - & - & -\\
        MSL~\cite{AAAI22MSL}& V-Swin & 85.30 & -  &  78.28 & -\\
        WSAL~\cite{TIP_WSL}&  I3D& 85.38&  67.38  & - & -\\
        ECU~\cite{CVPR23ECU}& V-Swin &  86.22 & -   &  - & -\\
        MGFN~\cite{AAAI23MGFN}&  V-Swin&  86.67 & -   & -  & - \\
        UMIL~\cite{CVPR23UMIL}& CLIP& 86.75 & 68.68 & - & -   \\
        TSA~\cite{ICIP2023CLIP}& CLIP& 87.58 &  -&  82.17& -  \\
	\midrule
    \bf GS-MoE (Ours) & \bf  I3D& \bf  91.58  & \bf 83.86 &   \bf 82.89 & \bf 85.74  \\
    & & {\color{red}\scriptsize(+3.56\%)} & {\color{red}\scriptsize(+13.63\%)} &  & \\
    \bottomrule
\end{tabular}
\caption{State-of-the-art comparisons on UCF-Crime and XD-Violence datasets. The best results are written in \textbf{bold}.}
\label{tab:testing}
\begin{tabular}{l|cccc}
    \toprule
    \textbf{Model} & \multicolumn{4}{c}{\textbf{MSAD}} \\
    & \textbf{AUC} & \textbf{AUC\textsubscript{A}}  & \textbf{AP} & \textbf{AP\textsubscript{A}} \\
     \midrule
    RTFM\citep{iccv21} & 86.65 & - & - & - \\
    MGFN\citep{AAAI23MGFN} & 84.96 & - & - & -   \\
    TEVAD\citep{chen2023tevad} & 86.82 & - & - & -  \\
    \midrule
    UR-DMU\citep{AAAI23URDMU} & 85.02 & -  & - & - \\
    UR-DMU {\color{red}*} & 85.78 & 67.95 & 67.35 & 75.30 \\
    \midrule
    \bf GS-MoE & \bf 87.72 & \bf69.54 & \bf 68.26 & \bf76.68\\
    \bottomrule
\end{tabular}
\caption{State-of-the-art comparisons on MSAD. {\color{red}*} indicates our own implementation. The best results are written in \textbf{bold}.}
\label{tab:testing_msad}
\end{table} 

\section{Experiments}
\textbf{Datasets.} We conduct our experiments on two widely-used Weakly-Supervised Video Anomaly Detection (WSVAD) datasets, namely UCF-Crime~\cite{cvpr18} and XD-Violence~\cite{eccv2020}. We further experiment on the recent MSAD dataset~\cite{msad2024}. Importantly, for all datasets, the training videos are annotated with only video-level labels, without access to frame-level annotations.

\textbf{Evaluation Metrics.}  
We adhere to the evaluation protocols established in prior works~\cite{CVPR23UMIL, AAAI2024_vadclip, cvpr18, eccv2020}. To ensure comprehensive evaluation, we utilize multiple indicators, such as frame-level Average Precision (AP), Abnormal AP (AP\textsubscript{A}) for XD-Violence and Area Under the Curve (AUC),  Abnormal AUC (AUC\textsubscript{A}) for UCF-Crime dataset. The AP and AUC metrics show the method robustness towards both normal and anomaly videos. However, AP\textsubscript{A} and AUC\textsubscript{A} allows to exclude normal videos where all snippets are labeled as normal and retain only the abnormal videos containing both normal and anomalous snippets. This poses a more meaningful challenge to the model's ability to accurately localize anomalies.

\begin{table}[ht]
\centering
\addtolength{\tabcolsep}{-0.45em}

\begin{tabular}{c|c|cc|cc|cc}
    \toprule
    \textbf{Baseline} & \textbf{TGS} & \multicolumn{2}{c|}{\textbf{MoE}} & \multicolumn{2}{c|}{\textbf{AUC(\%)}} & \multicolumn{2}{c}{\textbf{AP\textsubscript{A}(\%)}}\\
    & & \textbf{Experts} & \textbf{Gate} & \textbf{UCF} & \textbf{XD-V} & \textbf{UCF} & \textbf{XD-V} \\
    \midrule
    \cmark & - & - & - & 86.97 & 94.07 & 45.65 & 82.91 \\
    \cmark & \cmark & - & - & 87.84 & 94.13 & 46.01 & 83.39  \\
    \cmark & \cmark & \cmark & - & 89.53 & 94.29 & 47.17 & 84.16 \\
    \cmark & \cmark & \cmark & \cmark & \textbf{91.58} & \textbf{94.52} & \textbf{51.63} & \textbf{85.74}\\
    \bottomrule
\end{tabular}
\caption{ Impact of each component in GS-MoE framework on UCF-Crime and XD-Violence datasets.}
\label{tab:compres}
\end{table} 
\begin{table*}[ht]

\begin{center}
\scalebox{0.95}{
	\begin{tabular}{l!{\hspace{-.2cm}}|c!{\hspace{-.3cm}}c!{\hspace{-.3cm}}c!{\hspace{-.3cm}}c!{\hspace{-.3cm}}c!{\hspace{-.3cm}}c!{\hspace{-.3cm}}c!{\hspace{-.3cm}}c!{\hspace{-.3cm}}c!{\hspace{-.3cm}}c!{\hspace{-.3cm}}c!{\hspace{-.3cm}}c!{\hspace{-.3cm}}c!{\hspace{-.3cm}}}
		\hline
        \small \bf Expert	& \small \bf	Abuse	& \small	\bf Arrest	& \small	\bf Arson	& \small	\bf Assault	& \small \bf Burglary	& \small	\bf Explosion	& \small \bf	Fighting	& \small	\bf RoadAcc.	& \small \bf	Robbery	& \small \bf	Shooting	& \small \bf	Shoplifting	& \small \bf Stealing	& \small	\bf Vandalism	\\
            \midrule
        \small \bf  Mask & \small 50.02	&\small 50.51	&\small 49.27 &\small 50.72 &\small 49.49	&\small 49.92	&\small 49.95	&\small 49.91	&\small 50.04 &\small 49.20	&\small 49.39	&\small 50.52	&\small 49.87		\\
        \small \bf W/o Mask 	&\small \bf 86.37	&\small	\bf 55.48	&\small \bf 61.73	&\small	\bf 63.12	&\small	\bf 53.65	&\small \bf 57.04	&\small	\bf 65.14	&\small \bf 65.22	&\small \bf 72.37	&\small	\bf 60.89	&\small \bf 54.73	&\small	\bf 77.62	&\small	\bf 57.43		\\\hline
	\end{tabular}
    }
	\end{center}
    
\caption{Category-wise performance comparison on UCF-Crime dataset between the UR-DMU baseline model and GS-MoE without the expert model for a given class. Masking the relevant experts results in an almost random output from the gate model.}
\label{class_wise_Explainability}
\end{table*}
\begin{figure}
\begin{center}
\includegraphics[width=1\linewidth]{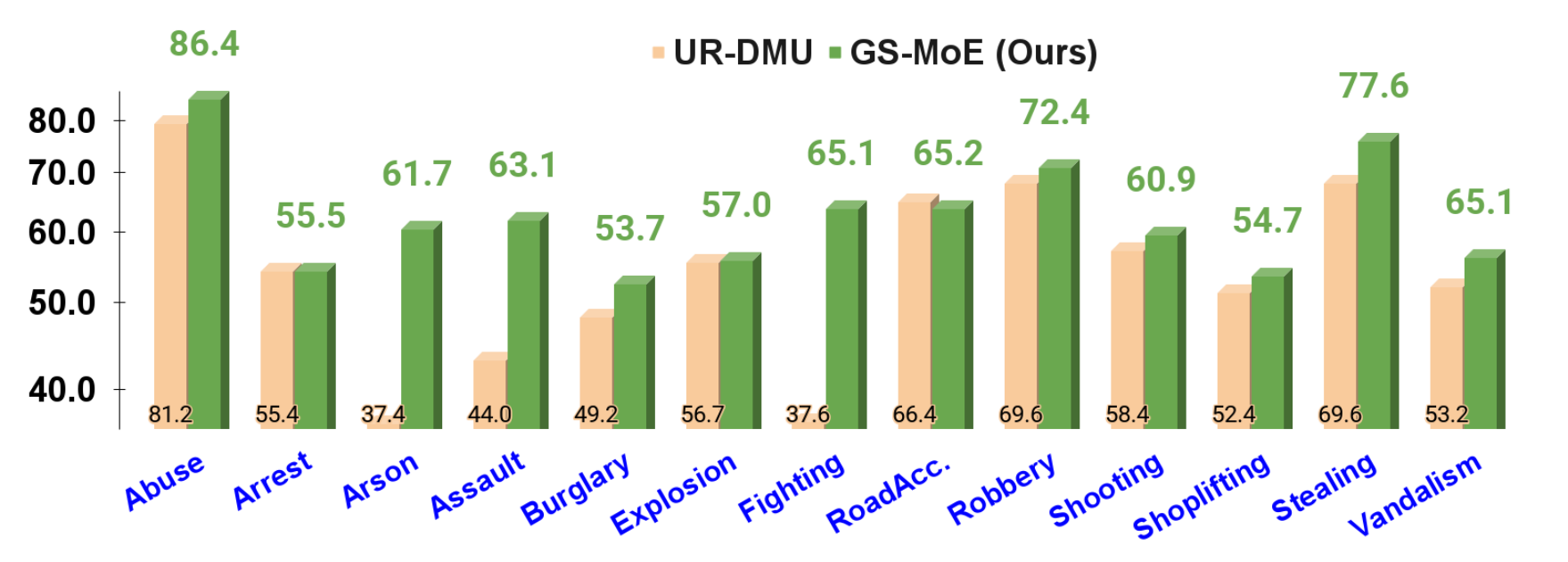}
\caption{Category-wise performance comparison with UR-DMU.}
\label{fig:categorywise}
\end{center}
\end{figure}

\begin{figure*}[ht]
\begin{subfigure}[ht]{0.5\textwidth}
    \includegraphics[width=\textwidth]{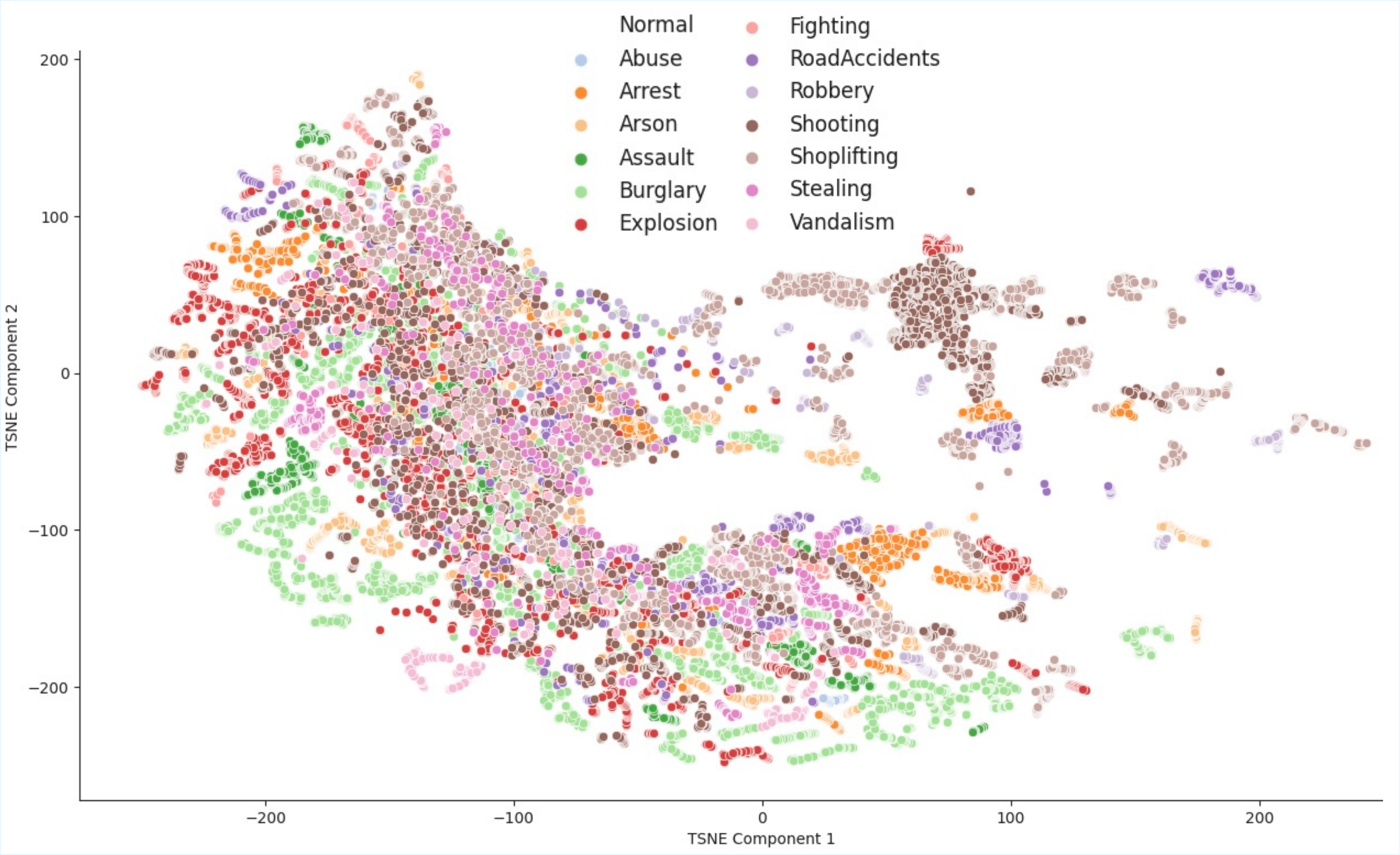}
    \caption{UR-DMU features.}
    \label{fig:urdmu}
\end{subfigure}
\begin{subfigure}[ht]{0.5\textwidth}
    \includegraphics[width=\textwidth]{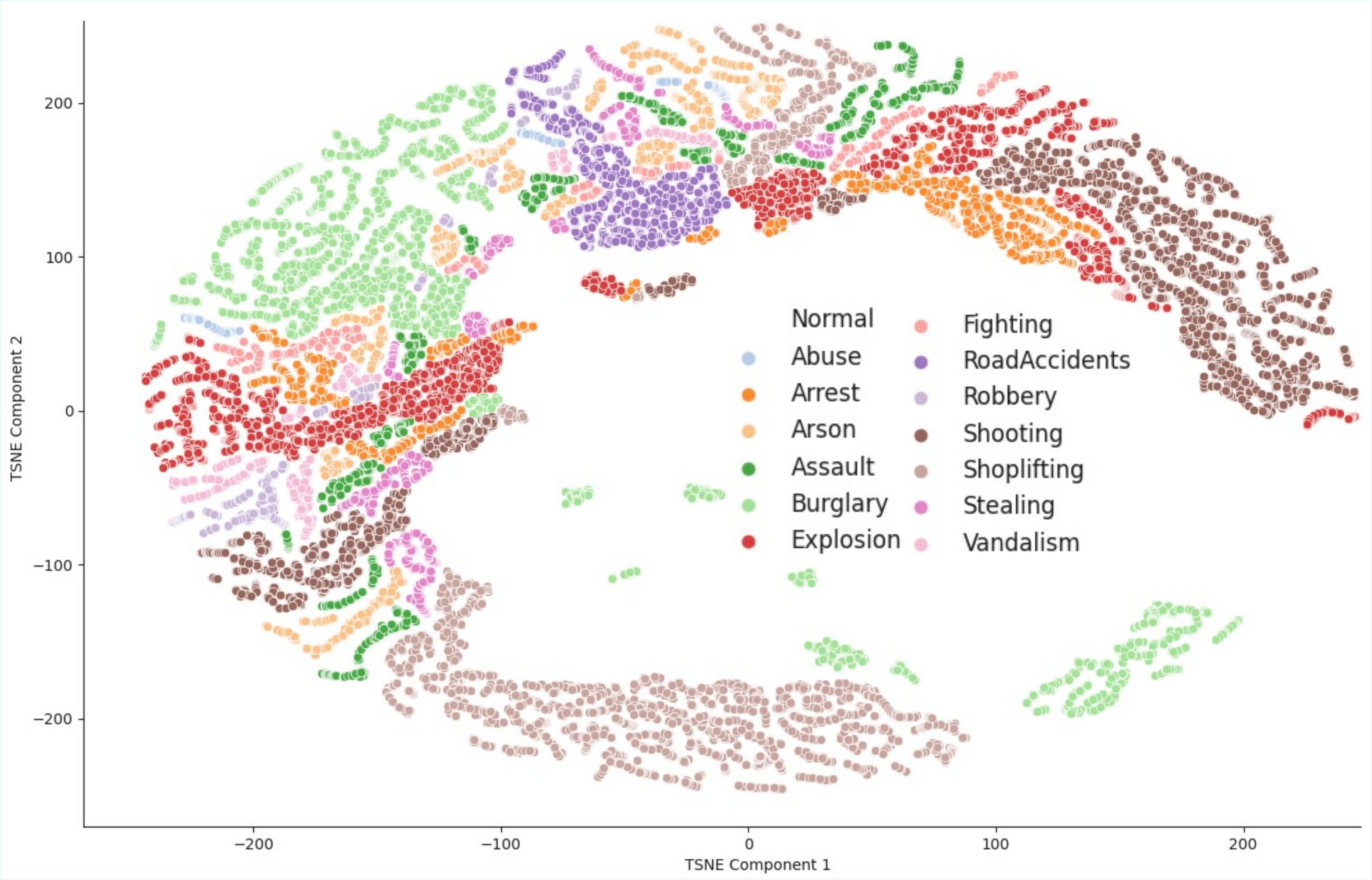}
    \caption{GS-MoE features.}
    \label{fig:tsne_experts}
\end{subfigure}
\caption{Category-wise t-SNE feature distribution comparison between the baseline, the experts and the gate model. }
\label{fig:tsne}
\end{figure*}

\begin{figure*}
\begin{center}
\includegraphics[width=1.\textwidth]{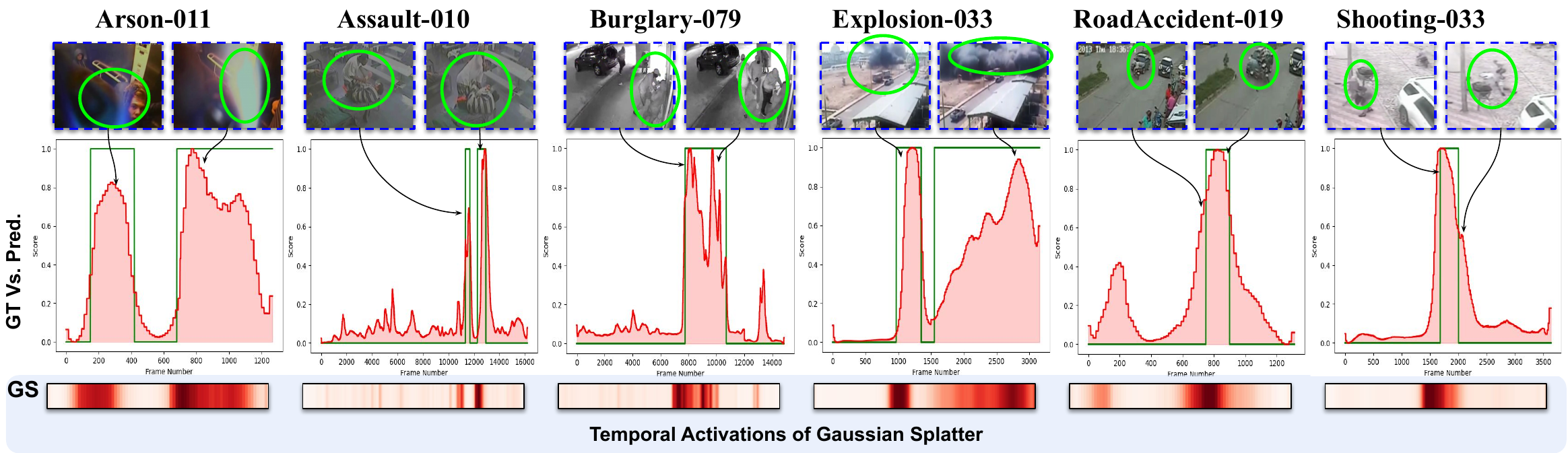}
\caption{Visualization of sample frames and ground truth (green shed) vs. prediction scores (red shed) for various cases in Row-1 and Row-2. For each plot in Row-2, the X and Y axis denotes the number of frames and corresponding anomaly scores. Row-3 shows the temporal activation (heatmaps) learned by Gaussian splatter (GS).}
\label{fig:splat}
\end{center}
\end{figure*}


\subsection{State-of-the-art Comparison} 
In our experiments, the proposed GS-MoE model outperforms prior state-of-the-art (SoTA) approaches across multiple metrics, as summarized in Table~\ref{tab:testing}. On the challenging \textbf{UCF-Crime dataset}, GS-MoE achieves an AUC of 91.58\%, surpassing the previous best model, VadCLIP~\cite{AAAI2024_vadclip}, by 3.56\%. This significant improvement illustrates the effectiveness of our model in detecting complex video anomalies in real-world datasets. Additionally, when considering the performance on the abnormal videos (AUC\textsubscript{A}) only, GS-MoE achieves a score of 83.86\%, which constitutes a remarkable 13.63\% improvement over the second-best approach, UR-DMU~\cite{AAAI23URDMU}, at 70.81\%. This result supports one key hypothesis of our work: different types of anomalies require class-specific fine-representations for more effective detection. UR-DMU performance remains limited due to feature-magnitude based optimization which overlooks the subtle cues and enhances the sharp cues. However, the proposed TGS loss promotes both subtle and sharp cues to take part in the separability optimization. Further, the mixture-of-experts architecture is capable of capturing these class-specific representations, leading to substantial performance gains, especially on complex anomalies.\\
On the \textbf{XD-Violence} dataset, GS-MoE achieves an AP score of 82.89\%, which is competitive with the best-performing  TSA~\cite{ICIP2023CLIP} model (82.89\%). Moreover, when focusing on anomalous videos only, GS-MoE achieves an AP\textsubscript{A} score of 85.74\%, outperforming the second-best approach, UR-DMU~\cite{AAAI23URDMU}, which achieved an AP\textsubscript{A} score of 83.94\%. Since the AP metric considers both normal and anomaly videos for evaluation, the performance gets elevated by accurately predicting many normal videos. 

As a result, methods performing well on the AP metric may still struggle in anomaly detection. 
The proposed method outperforms previous SoTA in the AP\textsubscript{A} metric, reinforcing its utility in real-world scenarios.
On the recently released \textbf{MSAD}, GS-MoE surpasses the available baseline models by up to 2.74\% on the AUC metric and establishes a new SoTA on the dataset. We report the other metrics as well, in order to provide a baseline for subsequent works.
\paragraph{Category-Wise Performance Analysis:} To bring additional analytical insights on the complex anomaly performance, Figure~\ref{fig:categorywise} provides an anomaly category-wise performance comparison between GS-MoE and the baseline UR-DMU method on the UCF-Crime dataset. Notably, significant performance boosts are recorded for complex categories like ``Arson", ``Assault", ``Fighting", ``Stealing" and ``Burglary", up to $+24.3\%$. These performance gains corroborate the benefits of GS-MoE in detecting complex video anomalies. Figure~\ref{fig:tsne} shows the t-SNE plot~\cite{JMLR:v9:vandermaaten08a} of the logits obtained at the first and third stages of GS-MoE for the anomalous videos in the test set. The plot in Figure~\ref{fig:urdmu}, obtained from the baseline UR-DMU, shows a low degree of separability. The class diversification performed by the experts and shown in Figure~\ref{fig:tsne_experts} demonstrates the capability of GS-MoE to learn enhanced class representations.

\subsection{Qualitative Results}
 As shown in Figure~\ref{fig:splat}, the Gaussian kernels extracted from the abnormal score contain a precise representation of the anomalous events present in videos of the UCF-Crime dataset. The kernel temporal activation (heatmaps) demonstrate the capabilities of this approach. By correctly distinguishing the peaks of the anomalous events and from the spurious peaks, the model is trained to predict high anomaly scores for the associated anomalous snippets. In the ``Assault-010" video sample, two peaks are detected in the abnormal score and the TGS finds a small variance for both, leading to a steep normal distribution for each of them. On the other hand, in the ``Arson-011" and ``Explosion-033" samples, the TGS creates much longer distributions by leading the model to estimate a large variance and producing a long time-window for the anomaly.

\subsection{Ablation Studies}

\paragraph{Component Impact:} Extensive ablation studies are conducted to evaluate the impact of each contribution to the final performance of GS-MoE, as shown in Table~\ref{tab:compres}. Fine-tuning the baseline UR-DMU model with the TGS loss in Equation~\ref{tgs} leads to a performance increase of $+1.77\%$ on the AUC metric of UCF-Crime, while the AP\textsubscript{A} of XD-Violence increases by $+0.48\%$. These results show that the new formulation of the WSVAD task is beneficial to existing methods as well.
The class-experts outperform the fine-tuned baseline by $+0.79\%$ on UCF-Crime. Notably, the AP\textsubscript{A} increases on both datasets, leading to $+1.16\%$ for UCF-Crime and +0.76\% on XD-Violence, further supporting the idea that different classes of anomaly should be treated separately. 
Adding the gate model to the framework brings the largest performance increment. For UCF-Crime, the AUC increases by $+2.05\%$ and the AP\textsubscript{A} by $+4.46\%$. On XD-Violence, we observe relatively smaller improvements, increasing AUC by $+0.23\%$ and AP\textsubscript{A} by $+1.68\%$. 

\begin{table}[ht]
\begin{center}
    \scalebox{0.8}{ 
    \begin{tabular}{c|c|c}
        \toprule
        \textbf{Datasets} & \textbf{W. TA features} & \textbf{W/o. TA features} \\
        \midrule
        \textbf{UCF-Crime} (\textit{AUC}) & \textbf{91.58} & 90.98 \\
        \textbf{XD-Violence} (\textit{AP\textsubscript{A}}) & \textbf{85.74}& 81.45 \\
        \bottomrule
    \end{tabular}
    }
\end{center}
\caption{Evaluation of the importance of the task-aware \textbf{(TA)} features for the gate model on the key metrics of the UCF-Crime and XD-Violence datasets.}
\label{Abl_1b}
\end{table}
\paragraph{Task-Aware Features:} In order to further analyze this performance increment, the gate model was trained with and without the task-aware features. The results of this experiment are shown in Table~\ref{Abl_1b}. The task-aware features seem to have a key role in the performance on the AP\textsubscript{A} metric of XD-Violence. In fact, the Gate model trained with the task-aware features outperforms the other configuration by 4.29\% on this setting, and by 0.6\% on UCF-Crime. 
\paragraph{Class-Experts Impact:} The relevance of the expert models on the performance of the gate model is measured with the class-wise AUC score obtained by masking the respective class expert on the UCF-Crime dataset. The results of this experiment are shown in Table~\ref{class_wise_Explainability}. By masking the experts, the measured AUC hovers around 50\% for each class. On the other hand, the gate model predictions are much improved when the relevant expert score is included, leading to a significant performance boost. 
\paragraph{Class experts vs cluster experts:} In practical applications, anomalies often span multiple classes, making it challenging to train a predefined set of specialized experts. To address this issue, we trained GS-MoE using cluster-based experts rather than class-specific experts. To form the data clusters, we calculated the average task-aware features for each anomalous video in the UCF-Crime training set and applied the K-Means algorithm~\citep{lloyd1982least} to group them. Each expert was then trained using videos from a single cluster combined with normal videos, resulting in $k$ specialized expert models. This approach enabled us to evaluate the model’s performance in real-world scenarios where the number of classes is undefined. The results are reported in Table~\ref{tab:clusters}. In this setting, GS-MoE is able to outperform current SoTA models by 0.56\% clustering the anomalous training videos in 7 clusters and using 7 experts while performing on par with other SoTA models using fewer experts. These results highlight the capabilities of GS-MoE in a real-world use case where the number of anomalous events is not predefined.

\begin{table}[ht]
    \centering
    \begin{tabular}{c|c}
    \toprule
        Model & AUC \\
        \midrule
       UR-DMU~\cite{AAAI23URDMU} & 86.97\\
        TSA~\cite{ICIP2023CLIP} & 87.58 \\
        TPWNG~\cite{CVPR24TPWNG} & 87.79 \\
        VadCLIP~\citep{AAAI2024_vadclip} & 88.02\\
        \midrule
        GS-MoE (5 clusters / 5 experts) & 87.35\\
        GS-MoE (6 clusters / 6 experts) & 88.03\\
        GS-MoE (7 clusters / 7 experts) & 88.58 \\
        \midrule
        \bf GS-MoE (class experts) & \bf 91.58 \\
        \bottomrule
    \end{tabular}
    \caption{Comparison between the performance of GS-MoE with varying number of experts.}
    \label{tab:clusters}
\end{table}

\section{Conclusion}
We proposed GS-MoE to provide a novel formulation for weakly-supervised video anomaly detection by leveraging Temporal Gaussian Splatting to overcome the limitations of previous methods. More specifically, we address the over-dependency on the most abnormal snippets for separability optimization. Our framework utilizes a mixture-of-experts architecture that learns category-specific fine-grained representations. building a correlation between coarse abnormal cues and fine-grained cues to learn a more compact representation for each category. Extensive experiments on challenging datasets across various metrics show GS-MoE consistently outperforms SoTA methods  with significant performance gains. In the future, we aim to leverage LLMs to provide more explainability to anomaly classes.

\paragraph{Acknowledgements:} This work was supported by Toyota Motor Europe (TME) and the French government, through the 3IA Cote d’Azur Investments managed by the National Research Agency (ANR) with the reference number ANR-19-P3IA-0002.

{
    \small
    \bibliographystyle{ieeenat_fullname}
    \bibliography{main}

\begin{thebibliography}{58}
\providecommand{\natexlab}[1]{#1}
\providecommand{\url}[1]{\texttt{#1}}
\expandafter\ifx\csname urlstyle\endcsname\relax
  \providecommand{\doi}[1]{doi: #1}\else
  \providecommand{\doi}{doi: \begingroup \urlstyle{rm}\Url}\fi

\bibitem[Acsintoae et~al.(2022)Acsintoae, Florescu, Georgescu, Mare, Sumedrea, Ionescu, Khan, and Shah]{Acsintoae_CVPR_2022}
Andra Acsintoae, Andrei Florescu, Mariana{-}Iuliana Georgescu, Tudor Mare, Paul Sumedrea, Radu~Tudor Ionescu, Fahad~Shahbaz Khan, and Mubarak Shah.
\newblock Ubnormal: New benchmark for supervised open-set video anomaly detection.
\newblock In \emph{Proceedings of the IEEE/CVF Conference on Computer Vision and Pattern Recognition (CVPR)}, 2022.

\bibitem[Bearman et~al.(2016)Bearman, Russakovsky, Ferrari, and Fei-Fei]{bearman2016s}
Amy Bearman, Olga Russakovsky, Vittorio Ferrari, and Li Fei-Fei.
\newblock What’s the point: Semantic segmentation with point supervision.
\newblock In \emph{European conference on computer vision}, pages 549--565. Springer, 2016.

\bibitem[Bertasius et~al.(2021)Bertasius, Wang, and Torresani]{bertasius2021space}
Gedas Bertasius, Heng Wang, and Lorenzo Torresani.
\newblock Is space-time attention all you need for video understanding?
\newblock In \emph{ICML}, page~4, 2021.

\bibitem[Carreira and Zisserman(2017)]{i3d}
Joao Carreira and Andrew Zisserman.
\newblock Quo vadis, action recognition? a new model and the kinetics dataset.
\newblock In \emph{The IEEE Conference on Computer Vision and Pattern Recognition (CVPR)}, 2017.

\bibitem[Chen et~al.(2021)Chen, Panda, Ramakrishnan, Feris, Cohn, Oliva, and Fan]{chen2021deep}
Chun-Fu~Richard Chen, Rameswar Panda, Kandan Ramakrishnan, Rogerio Feris, John Cohn, Aude Oliva, and Quanfu Fan.
\newblock Deep analysis of cnn-based spatio-temporal representations for action recognition.
\newblock In \emph{Proceedings of the IEEE/CVF conference on computer vision and pattern recognition}, pages 6165--6175, 2021.

\bibitem[Chen et~al.(2023{\natexlab{a}})Chen, Ma, Yew, Hur, and Khoo]{chen2023tevad}
Weiling Chen, Keng~Teck Ma, Zi~Jian Yew, Minhoe Hur, and David Aik-Aun Khoo.
\newblock Tevad: Improved video anomaly detection with captions.
\newblock In \emph{Proceedings of the IEEE/CVF Conference on Computer Vision and Pattern Recognition}, pages 5549--5559, 2023{\natexlab{a}}.

\bibitem[Chen et~al.(2023{\natexlab{b}})Chen, Liu, Zhang, Fok, Qi, and Wu]{AAAI23MGFN}
Yingxian Chen, Zhengzhe Liu, Baoheng Zhang, Wilton Fok, Xiaojuan Qi, and Yik-Chung Wu.
\newblock Mgfn: Magnitude-contrastive glance-and-focus network for weakly-supervised video anomaly detection.
\newblock In \emph{Proceedings of the AAAI Conference on Artificial Intelligence}, pages 387--395, 2023{\natexlab{b}}.

\bibitem[Cho et~al.(2023)Cho, Kim, Hwang, Park, Lee, and Lee]{CVPR23LAA}
MyeongAh Cho, Minjung Kim, Sangwon Hwang, Chaewon Park, Kyungjae Lee, and Sangyoun Lee.
\newblock Look around for anomalies: Weakly-supervised anomaly detection via context-motion relational learning.
\newblock In \emph{Proceedings of the IEEE/CVF Conference on Computer Vision and Pattern Recognition}, pages 12137--12146, 2023.

\bibitem[Eigen et~al.(2013)Eigen, Ranzato, and Sutskever]{eigen2013learning}
David Eigen, Marc'Aurelio Ranzato, and Ilya Sutskever.
\newblock Learning factored representations in a deep mixture of experts.
\newblock \emph{arXiv preprint arXiv:1312.4314}, 2013.

\bibitem[Fedus et~al.(2022)Fedus, Zoph, and Shazeer]{fedus2022switch}
William Fedus, Barret Zoph, and Noam Shazeer.
\newblock Switch transformers: Scaling to trillion parameter models with simple and efficient sparsity.
\newblock \emph{Journal of Machine Learning Research}, 23\penalty0 (120):\penalty0 1--39, 2022.

\bibitem[Feng et~al.(2021)Feng, Hong, and Zheng]{cvpr21}
Jia-Chang Feng, Fa-Ting Hong, and Wei-Shi Zheng.
\newblock Mist: Multiple instance self-training framework for video anomaly detection.
\newblock In \emph{Proceedings of the IEEE/CVF Conference on Computer Vision and Pattern Recognition}, pages 14009--14018, 2021.

\bibitem[Georgescu et~al.(2021{\natexlab{a}})Georgescu, Barbalau, Ionescu, Khan, Popescu, and Shah]{georgescu2021anomaly}
Mariana-Iuliana Georgescu, Antonio Barbalau, Radu~Tudor Ionescu, Fahad~Shahbaz Khan, Marius Popescu, and Mubarak Shah.
\newblock Anomaly detection in video via self-supervised and multi-task learning.
\newblock In \emph{Proceedings of the IEEE/CVF conference on computer vision and pattern recognition}, pages 12742--12752, 2021{\natexlab{a}}.

\bibitem[Georgescu et~al.(2021{\natexlab{b}})Georgescu, Ionescu, Khan, Popescu, and Shah]{georgescu2021background}
Mariana~Iuliana Georgescu, Radu~Tudor Ionescu, Fahad~Shahbaz Khan, Marius Popescu, and Mubarak Shah.
\newblock A background-agnostic framework with adversarial training for abnormal event detection in video.
\newblock \emph{IEEE transactions on pattern analysis and machine intelligence}, 44\penalty0 (9):\penalty0 4505--4523, 2021{\natexlab{b}}.

\bibitem[Hendrycks and Gimpel(2016)]{gelu}
Dan Hendrycks and Kevin Gimpel.
\newblock Bridging nonlinearities and stochastic regularizers with gaussian error linear units.
\newblock \emph{CoRR}, abs/1606.08415, 2016.

\bibitem[Jain et~al.(2024)Jain, Hegde, Kusupati, Nagrani, Buch, Jain, Arnab, and Paul]{jain2024mixture}
Gagan Jain, Nidhi Hegde, Aditya Kusupati, Arsha Nagrani, Shyamal Buch, Prateek Jain, Anurag Arnab, and Sujoy Paul.
\newblock Mixture of nested experts: Adaptive processing of visual tokens.
\newblock \emph{arXiv preprint arXiv:2407.19985}, 2024.

\bibitem[Joo et~al.(2023)Joo, Vo, Yamazaki, and Le]{ICIP2023CLIP}
Hyekang~Kevin Joo, Khoa Vo, Kashu Yamazaki, and Ngan Le.
\newblock Clip-tsa: Clip-assisted temporal self-attention for weakly-supervised video anomaly detection.
\newblock In \emph{2023 IEEE International Conference on Image Processing (ICIP)}, pages 3230--3234. IEEE, 2023.

\bibitem[Kerbl et~al.(2023)Kerbl, Kopanas, Leimk{\"u}hler, and Drettakis]{kerbl20233d}
Bernhard Kerbl, Georgios Kopanas, Thomas Leimk{\"u}hler, and George Drettakis.
\newblock 3d gaussian splatting for real-time radiance field rendering.
\newblock \emph{ACM Trans. Graph.}, 42\penalty0 (4):\penalty0 139--1, 2023.

\bibitem[Kopanas et~al.(2021)Kopanas, Philip, Leimk{\"u}hler, and Drettakis]{kopanas2021point}
Georgios Kopanas, Julien Philip, Thomas Leimk{\"u}hler, and George Drettakis.
\newblock Point-based neural rendering with per-view optimization.
\newblock In \emph{Computer Graphics Forum}, pages 29--43. Wiley Online Library, 2021.

\bibitem[Lepikhin et~al.(2020)Lepikhin, Lee, Xu, Chen, Firat, Huang, Krikun, Shazeer, and Chen]{lepikhin2020gshard}
Dmitry Lepikhin, HyoukJoong Lee, Yuanzhong Xu, Dehao Chen, Orhan Firat, Yanping Huang, Maxim Krikun, Noam Shazeer, and Zhifeng Chen.
\newblock Gshard: Scaling giant models with conditional computation and automatic sharding.
\newblock \emph{arXiv preprint arXiv:2006.16668}, 2020.

\bibitem[Li et~al.(2024{\natexlab{a}})Li, Huang, Lu, Duan, and Huang]{4DGS}
Deqi Li, Shi-Sheng Huang, Zhiyuan Lu, Xinran Duan, and Hua Huang.
\newblock St-4dgs: Spatial-temporally consistent 4d gaussian splatting for efficient dynamic scene rendering.
\newblock In \emph{ACM SIGGRAPH 2024 Conference Papers}, New York, NY, USA, 2024{\natexlab{a}}. Association for Computing Machinery.

\bibitem[Li et~al.(2022{\natexlab{a}})Li, Cai, Zeng, and Zhao]{ECCV2022Scale}
Guoqiu Li, Guanxiong Cai, Xingyu Zeng, and Rui Zhao.
\newblock Scale-aware spatio-temporal relation learning for video anomaly detection.
\newblock In \emph{European Conference on Computer Vision}, pages 333--350. Springer, 2022{\natexlab{a}}.

\bibitem[Li et~al.(2022{\natexlab{b}})Li, Liu, and Jiao]{AAAI22MSL}
Shuo Li, Fang Liu, and Licheng Jiao.
\newblock Self-training multi-sequence learning with transformer for weakly supervised video anomaly detection.
\newblock In \emph{Proceedings of the AAAI Conference on Artificial Intelligence}, pages 1395--1403, 2022{\natexlab{b}}.

\bibitem[Li et~al.(2022{\natexlab{c}})Li, Liu, and Jiao]{li2022self}
Shuo Li, Fang Liu, and Licheng Jiao.
\newblock Self-training multi-sequence learning with transformer for weakly supervised video anomaly detection.
\newblock In \emph{Proceedings of the AAAI Conference on Artificial Intelligence}, pages 1395--1403, 2022{\natexlab{c}}.

\bibitem[Li et~al.(2024{\natexlab{b}})Li, Chen, Li, and Xu]{li2024spacetime}
Zhan Li, Zhang Chen, Zhong Li, and Yi Xu.
\newblock Spacetime gaussian feature splatting for real-time dynamic view synthesis.
\newblock In \emph{Proceedings of the IEEE/CVF Conference on Computer Vision and Pattern Recognition}, pages 8508--8520, 2024{\natexlab{b}}.

\bibitem[Lloyd(1982)]{lloyd1982least}
Stuart Lloyd.
\newblock Least squares quantization in pcm.
\newblock \emph{IEEE transactions on information theory}, 28\penalty0 (2):\penalty0 129--137, 1982.

\bibitem[Loshchilov and Hutter(2017)]{adamw}
Ilya Loshchilov and Frank Hutter.
\newblock Fixing weight decay regularization in adam.
\newblock \emph{CoRR}, abs/1711.05101, 2017.

\bibitem[Lv et~al.(2021)Lv, Zhou, Cui, Xu, Li, and Yang]{TIP_WSL}
Hui Lv, Chuanwei Zhou, Zhen Cui, Chunyan Xu, Yong Li, and Jian Yang.
\newblock Localizing anomalies from weakly-labeled videos.
\newblock \emph{IEEE transactions on image processing}, 30:\penalty0 4505--4515, 2021.

\bibitem[Lv et~al.(2023)Lv, Yue, Sun, Luo, Cui, and Zhang]{CVPR23UMIL}
Hui Lv, Zhongqi Yue, Qianru Sun, Bin Luo, Zhen Cui, and Hanwang Zhang.
\newblock Unbiased multiple instance learning for weakly supervised video anomaly detection.
\newblock In \emph{Proceedings of the IEEE/CVF Conference on Computer Vision and Pattern Recognition}, pages 8022--8031, 2023.

\bibitem[Majhi et~al.(2024{\natexlab{a}})Majhi, Dai, Kong, Garattoni, Francesca, and Bremond]{HSN}
Snehashis Majhi, Rui Dai, Quan Kong, Lorenzo Garattoni, Gianpiero Francesca, and Francois Bremond.
\newblock Human-scene network: A novel baseline with self-rectifying loss for weakly supervised video anomaly detection.
\newblock \emph{Computer Vision and Image Understanding}, 241:\penalty0 103955, 2024{\natexlab{a}}.

\bibitem[Majhi et~al.(2024{\natexlab{b}})Majhi, Dai, Kong, Garattoni, Francesca, and Br{\'e}mond]{majhi2024oe}
Snehashis Majhi, Rui Dai, Quan Kong, Lorenzo Garattoni, Gianpiero Francesca, and Fran{\c{c}}ois Br{\'e}mond.
\newblock Oe-ctst: Outlier-embedded cross temporal scale transformer for weakly-supervised video anomaly detection.
\newblock In \emph{Proceedings of the IEEE/CVF winter conference on applications of computer vision}, pages 8574--8583, 2024{\natexlab{b}}.

\bibitem[Majhi et~al.(2025)Majhi, D'Amicantonio, Dantcheva, Kong, Garattoni, Francesca, Bondarev, and Bremond]{Majhi_2025_CVPR}
Snehashis Majhi, Giacomo D'Amicantonio, Antitza Dantcheva, Quan Kong, Lorenzo Garattoni, Gianpiero Francesca, Egor Bondarev, and Francois Bremond.
\newblock Just dance with pi! a poly-modal inductor for weakly-supervised video anomaly detection.
\newblock In \emph{Proceedings of the IEEE/CVF Conference on Computer Vision and Pattern Recognition (CVPR)}, pages 24265--24274, 2025.

\bibitem[Mustafa et~al.(2022)Mustafa, Riquelme, Puigcerver, Jenatton, and Houlsby]{mustafa2022multimodal}
Basil Mustafa, Carlos Riquelme, Joan Puigcerver, Rodolphe Jenatton, and Neil Houlsby.
\newblock Multimodal contrastive learning with limoe: the language-image mixture of experts.
\newblock \emph{Advances in Neural Information Processing Systems}, 35:\penalty0 9564--9576, 2022.

\bibitem[Puigcerver et~al.(2024)Puigcerver, Riquelme, Mustafa, and Houlsby]{puigcerver2024sparsesoftmixturesexperts}
Joan Puigcerver, Carlos Riquelme, Basil Mustafa, and Neil Houlsby.
\newblock From sparse to soft mixtures of experts, 2024.

\bibitem[Purwanto et~al.(2021)Purwanto, Chen, and Fang]{iccv21_Dance}
Didik Purwanto, Yie-Tarng Chen, and Wen-Hsien Fang.
\newblock Dance with self-attention: A new look of conditional random fields on anomaly detection in videos.
\newblock In \emph{Proceedings of the IEEE/CVF International Conference on Computer Vision}, pages 173--183, 2021.

\bibitem[Riquelme et~al.(2021)Riquelme, Puigcerver, Mustafa, Neumann, Jenatton, Susano~Pinto, Keysers, and Houlsby]{riquelme2021scaling}
Carlos Riquelme, Joan Puigcerver, Basil Mustafa, Maxim Neumann, Rodolphe Jenatton, Andr{\'e} Susano~Pinto, Daniel Keysers, and Neil Houlsby.
\newblock Scaling vision with sparse mixture of experts.
\newblock \emph{Advances in Neural Information Processing Systems}, 34:\penalty0 8583--8595, 2021.

\bibitem[Sultani et~al.(2018{\natexlab{a}})Sultani, Chen, and Shah]{cvpr18}
Waqas Sultani, Chen Chen, and Mubarak Shah.
\newblock Real-world anomaly detection in surveillance videos.
\newblock In \emph{Proceedings of the IEEE Conference on Computer Vision and Pattern Recognition}, pages 6479--6488, 2018{\natexlab{a}}.

\bibitem[Sultani et~al.(2018{\natexlab{b}})Sultani, Chen, and Shah]{sultani2018real}
Waqas Sultani, Chen Chen, and Mubarak Shah.
\newblock Real-world anomaly detection in surveillance videos.
\newblock In \emph{Proceedings of the IEEE conference on computer vision and pattern recognition}, pages 6479--6488, 2018{\natexlab{b}}.

\bibitem[Tian et~al.(2021{\natexlab{a}})Tian, Pang, Chen, Singh, Verjans, and Carneiro]{iccv21}
Yu Tian, Guansong Pang, Yuanhong Chen, Rajvinder Singh, Johan~W Verjans, and Gustavo Carneiro.
\newblock Weakly-supervised video anomaly detection with robust temporal feature magnitude learning.
\newblock In \emph{Proceedings of the IEEE/CVF International Conference on Computer Vision}, pages 4975--4986, 2021{\natexlab{a}}.

\bibitem[Tian et~al.(2021{\natexlab{b}})Tian, Pang, Chen, Singh, Verjans, and Carneiro]{tian2021weakly}
Yu Tian, Guansong Pang, Yuanhong Chen, Rajvinder Singh, Johan~W Verjans, and Gustavo Carneiro.
\newblock Weakly-supervised video anomaly detection with robust temporal feature magnitude learning.
\newblock In \emph{Proceedings of the IEEE/CVF international conference on computer vision}, pages 4975--4986, 2021{\natexlab{b}}.

\bibitem[van~der Maaten and Hinton(2008)]{JMLR:v9:vandermaaten08a}
Laurens van~der Maaten and Geoffrey Hinton.
\newblock Visualizing data using t-sne.
\newblock \emph{Journal of Machine Learning Research}, 9\penalty0 (86):\penalty0 2579--2605, 2008.

\bibitem[Wu et~al.(2020)Wu, Liu, Shi, Sun, Shao, Wu, and Yang]{eccv2020}
Peng Wu, Jing Liu, Yujia Shi, Yujia Sun, Fangtao Shao, Zhaoyang Wu, and Zhiwei Yang.
\newblock Not only look, but also listen: Learning multimodal violence detection under weak supervision.
\newblock In \emph{European Conference on Computer Vision}, pages 322--339. Springer, 2020.

\bibitem[Wu et~al.(2022)Wu, Liu, and Liu]{wu2022weakly}
Peng Wu, Xiaotao Liu, and Jing Liu.
\newblock Weakly supervised audio-visual violence detection.
\newblock \emph{IEEE Transactions on Multimedia}, 25:\penalty0 1674--1685, 2022.

\bibitem[Wu et~al.(2024)Wu, Zhou, Pang, Zhou, Yan, Wang, and Zhang]{AAAI2024_vadclip}
Peng Wu, Xuerong Zhou, Guansong Pang, Lingru Zhou, Qingsen Yan, Peng Wang, and Yanning Zhang.
\newblock Vadclip: Adapting vision-language models for weakly supervised video anomaly detection.
\newblock In \emph{Proceedings of the AAAI Conference on Artificial Intelligence}, pages 6074--6082, 2024.

\bibitem[Yan et~al.(2023)Yan, Hu, Sun, Tang, Zhu, Dong, Van~Gool, and Zhang]{yan2023towards}
Qingsen Yan, Tao Hu, Yuan Sun, Hao Tang, Yu Zhu, Wei Dong, Luc Van~Gool, and Yanning Zhang.
\newblock Towards high-quality hdr deghosting with conditional diffusion models.
\newblock \emph{IEEE Transactions on Circuits and Systems for Video Technology}, 2023.

\bibitem[Yang et~al.(2024)Yang, Liu, and Wu]{CVPR24TPWNG}
Zhiwei Yang, Jing Liu, and Peng Wu.
\newblock Text prompt with normality guidance for weakly supervised video anomaly detection.
\newblock In \emph{Proceedings of the IEEE/CVF Conference on Computer Vision and Pattern Recognition (CVPR)}, pages 18899--18908, 2024.

\bibitem[Yu et~al.(2020)Yu, Wang, Cai, Zhu, Xu, Yin, and Kloft]{yu2020cloze}
Guang Yu, Siqi Wang, Zhiping Cai, En Zhu, Chuanfu Xu, Jianping Yin, and Marius Kloft.
\newblock Cloze test helps: Effective video anomaly detection via learning to complete video events.
\newblock In \emph{Proceedings of the 28th ACM international conference on multimedia}, pages 583--591, 2020.

\bibitem[Yu et~al.(2022)Yu, Liu, Cheng, Feng, and Zhang]{yu2022modality}
Jiashuo Yu, Jinyu Liu, Ying Cheng, Rui Feng, and Yuejie Zhang.
\newblock Modality-aware contrastive instance learning with self-distillation for weakly-supervised audio-visual violence detection.
\newblock In \emph{Proceedings of the 30th ACM international conference on multimedia}, pages 6278--6287, 2022.

\bibitem[Zhang et~al.(2023{\natexlab{a}})Zhang, Li, Qi, Wang, Qing, Huang, and Yang]{CVPR23ECU}
Chen Zhang, Guorong Li, Yuankai Qi, Shuhui Wang, Laiyun Qing, Qingming Huang, and Ming-Hsuan Yang.
\newblock Exploiting completeness and uncertainty of pseudo labels for weakly supervised video anomaly detection.
\newblock In \emph{Proceedings of the IEEE/CVF Conference on Computer Vision and Pattern Recognition}, pages 16271--16280, 2023{\natexlab{a}}.

\bibitem[Zhang et~al.(2023{\natexlab{b}})Zhang, Li, Qi, Wang, Qing, Huang, and Yang]{zhang2023exploiting}
Chen Zhang, Guorong Li, Yuankai Qi, Shuhui Wang, Laiyun Qing, Qingming Huang, and Ming-Hsuan Yang.
\newblock Exploiting completeness and uncertainty of pseudo labels for weakly supervised video anomaly detection.
\newblock In \emph{Proceedings of the IEEE/CVF Conference on Computer Vision and Pattern Recognition}, pages 16271--16280, 2023{\natexlab{b}}.

\bibitem[Zhang et~al.(2024{\natexlab{a}})Zhang, Wang, Xu, Huang, Han, Wang, Gao, Zhang, and Sang]{zhang2024glancevad}
Huaxin Zhang, Xiang Wang, Xiaohao Xu, Xiaonan Huang, Chuchu Han, Yuehuan Wang, Changxin Gao, Shanjun Zhang, and Nong Sang.
\newblock Glancevad: Exploring glance supervision for label-efficient video anomaly detection.
\newblock \emph{arXiv preprint arXiv:2403.06154}, 2024{\natexlab{a}}.

\bibitem[Zhang et~al.(2019)Zhang, Qing, and Miao]{icip2019}
Jiangong Zhang, Laiyun Qing, and Jun Miao.
\newblock Temporal convolutional network with complementary inner bag loss for weakly supervised anomaly detection.
\newblock In \emph{2019 IEEE International Conference on Image Processing (ICIP)}, pages 4030--4034. IEEE, 2019.

\bibitem[Zhang et~al.(2024{\natexlab{b}})Zhang, Zhao, Zhou, Wu, Zheng, Wang, and Liu]{zhang2024togsgaussiansplattingtemporal}
Shuai Zhang, Huangxuan Zhao, Zhenghong Zhou, Guanjun Wu, Chuansheng Zheng, Xinggang Wang, and Wenyu Liu.
\newblock Togs: Gaussian splatting with temporal opacity offset for real-time 4d dsa rendering, 2024{\natexlab{b}}.

\bibitem[Zhong et~al.(2019{\natexlab{a}})Zhong, Li, Kong, Liu, Li, and Li]{cvpr19}
Jia-Xing Zhong, Nannan Li, Weijie Kong, Shan Liu, Thomas~H. Li, and Ge Li.
\newblock Graph convolutional label noise cleaner: Train a plug-and-play action classifier for anomaly detection.
\newblock In \emph{The IEEE Conference on Computer Vision and Pattern Recognition (CVPR)}, 2019{\natexlab{a}}.

\bibitem[Zhong et~al.(2019{\natexlab{b}})Zhong, Li, Kong, Liu, Li, and Li]{zhong2019graph}
Jia-Xing Zhong, Nannan Li, Weijie Kong, Shan Liu, Thomas~H Li, and Ge Li.
\newblock Graph convolutional label noise cleaner: Train a plug-and-play action classifier for anomaly detection.
\newblock In \emph{Proceedings of the IEEE/CVF conference on computer vision and pattern recognition}, pages 1237--1246, 2019{\natexlab{b}}.

\bibitem[Zhou et~al.(2023{\natexlab{a}})Zhou, Yu, and Yang]{AAAI23URDMU}
Hang Zhou, Junqing Yu, and Wei Yang.
\newblock Dual memory units with uncertainty regulation for weakly supervised video anomaly detection.
\newblock \emph{arXiv preprint arXiv:2302.05160}, 2023{\natexlab{a}}.

\bibitem[Zhou et~al.(2023{\natexlab{b}})Zhou, Yu, and Yang]{zhou2023dual}
Hang Zhou, Junqing Yu, and Wei Yang.
\newblock Dual memory units with uncertainty regulation for weakly supervised video anomaly detection.
\newblock In \emph{Proceedings of the AAAI Conference on Artificial Intelligence}, pages 3769--3777, 2023{\natexlab{b}}.

\bibitem[Zhu et~al.(2024)Zhu, Wang, Raj, Gedeon, and Chen]{msad2024}
Liyun Zhu, Lei Wang, Arjun Raj, Tom Gedeon, and Chen Chen.
\newblock Advancing video anomaly detection: A concise review and a new dataset.
\newblock In \emph{The Thirty-eighth Conference on Neural Information Processing Systems Datasets and Benchmarks Track}, 2024.

\bibitem[Zhu and Newsam(2019)]{bmvc19}
Yi Zhu and Shawn Newsam.
\newblock Motion-aware feature for improved video anomaly detection.
\newblock \emph{arXiv preprint arXiv:1907.10211}, 2019.

\end{thebibliography}
}

\clearpage

\twocolumn[
\begin{@twocolumnfalse}

\begin{center}
    \textbf{\Large Mixture of Experts Guided by Gaussian Splatters Matters: A new Approach to Weakly-Supervised Video Anomaly Detection}
    \vspace{1.5em}
    
    {\Large \bfseries Supplementary Material}
\end{center}

\vspace{2em} 

\setcounter{page}{1}
\setcounter{section}{0}
\setcounter{figure}{0}
\setcounter{table}{0} 

\renewcommand{\thepage}{S\arabic{page}}
\renewcommand{\thesection}{S\arabic{section}}
\renewcommand{\thefigure}{S\arabic{figure}}
\renewcommand{\thetable}{S\arabic{table}}

\end{@twocolumnfalse}
]

We include additional details and results about GS-MoE. In Section~\ref{sec:implementation}, we report the details on the training of the models presented in the main paper as well as the data pre-processing. Section~\ref{sec:tgs} presents an additional ablation study on the peak-detection mechanism, while Section~\ref{sec:soft} includes the design and experimental results on an alternative soft-MoE architecture implemented for GS-MoE. Section~\ref{sec:costs} reports the computational costs of the proposed framework. In Section~\ref{sec:qa}, we include a qualitative analysis of the most common failure cases of GS-MoE and finally, Section~\ref{sec:ubnormal} contains the experimental results obtained on the UBnormal~\citep{Acsintoae_CVPR_2022} dataset.

\section{Implementation Details}
\label{sec:implementation}
The video features were obtained with the I3D model~\cite{i3d} pre-trained on Kinetics-400 with sliding windows of 16 frames. The I3D implementation chosen is the ResNet50, which is proven to be one of the best-performing~\cite{chen2021deep}. The transformer blocks implemented in the experts and gate model do not have positional embeddings and class tokens.
The I3D features of each video have dimensions 1xNx1024, where N is the number of snippets in the video. Each snippet contains 16 consecutive frames. Each video has a different number of snippets. In order to create batches of videos, the snippets of each video are linearly projected to a fixed dimension $D$. To do so, the snippets are evenly spaced over $D$. Following the common practice in the WSVAD field, $D$ is set to 200. For example, if a video contains 100 snippets, they are projected as $[1, 1.5, 2, 2.5, ... , 99, 99.5, 100]$, where the decimal values indicate that the respective projected snippet is the weighted average between the previous snippet and the following snippet. This is done only for the training set videos, for the testing set it is not necessary to create batches. Therefore, the features dimension used as input for the expert models have dimension Bx200x1024, where B is the batch size.
All models were implemented in PyTorch and trained on a single NVIDIA RTX A4500 GPU. The models were trained using the AdamW~\cite{adamw} optimizer. The batch size was set at 128, containing 64 normal and 64 abnormal videos. Under these conditions, the entire training procedure requires about three hours, while testing on the UCF-Crime test set requires 55 seconds. For training stability, during the first epoch, the models are trained with the $L_{topk-norm}$ component of the following equation:
\begin{equation}
    L_{TGS} = L_{topk-norm} + BCE(y, \hat{y})
\label{tgs}
\end{equation}
For the same purposes, we employ the same smoothness and sparsity loss components as presented in~\cite{sultani2018real}. 

\textbf{Expert Model:} The expert models are composed of a transformer block and an MLP. The input of each expert model is the 1024-dimensional logits of the task-aware encoder. The transformer block first applies layer normalization to them, followed by a 2-head self-attention layer. The output is then added to the task-aware logits and further normalized via layer normalization. The resulting tensor is projected to a 512-dimensional space and a Relu activation is applied to introduce nonlinearities in the latent space. The features obtained this way are then projected back to a 1024-dimensional space and added to the output of the self-attention layer. The MLP is composed of four linear layers than progressively reduce the dimensionality of the transformer's output to 256, 128, 64 and finally 1, which is the expert's score. A Gelu activation function is applied between the second-to-last and the last linear layer. To ensure that the score is between 0 and 1, the sigmoid function is applied to the last layer's output. An expert model contains approximately 500 thousand parameters.

\textbf{Gate Model:} The expert scores are concatenated along the last dimension, creating a tensor of dimension Bx200xN. of experts. The tensor is projected to a 1024-dimensional space via a linear layer. Then, the bi-directional cross-attention layer is applied. In one direction, it takes the projected scores as values and the task-aware logits as key and queries. In the other direction, the task-aware logits are the values and the projected score are the keys and queries. The outputs of the bi-directional attention are concatenated along the last dimension, creating a tensor of dimension Bx200x2048. This is then fed to a transformer block similar to the one described for the expert models, with 4 attention head instead of 2. This difference is due to the fact that the input of the gate's transformer block is double the dimension of the input of the expert's transformer block. The MLP component of the gate model has the same architecture as the expert's MLP. The gate model contains approximately 1 million parameters.

\begin{figure*}[ht]
    \centering
    \includegraphics[width=\linewidth]{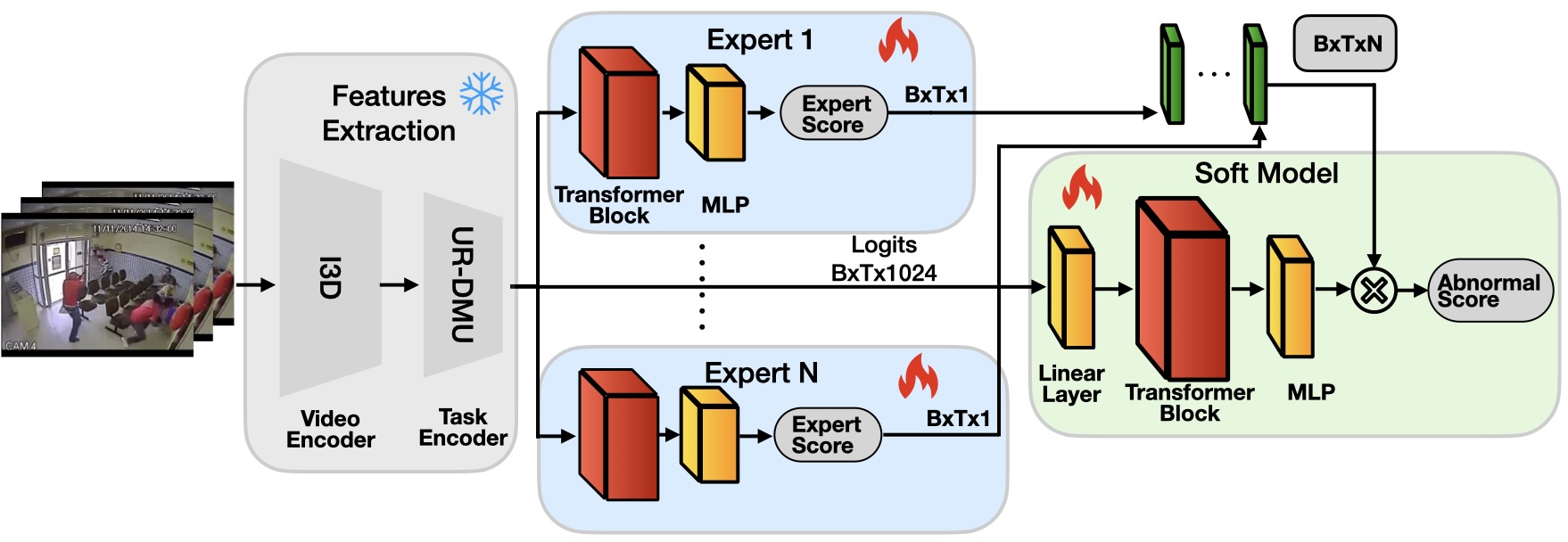}
    \caption{The soft-MoE architecture uses the scores estimated by the experts to inform the prediction made by the gate model.}
    \label{fig:soft_moe}
\end{figure*}

\begin{figure}[ht]
    \includegraphics[height=3.5cm,width=\linewidth]{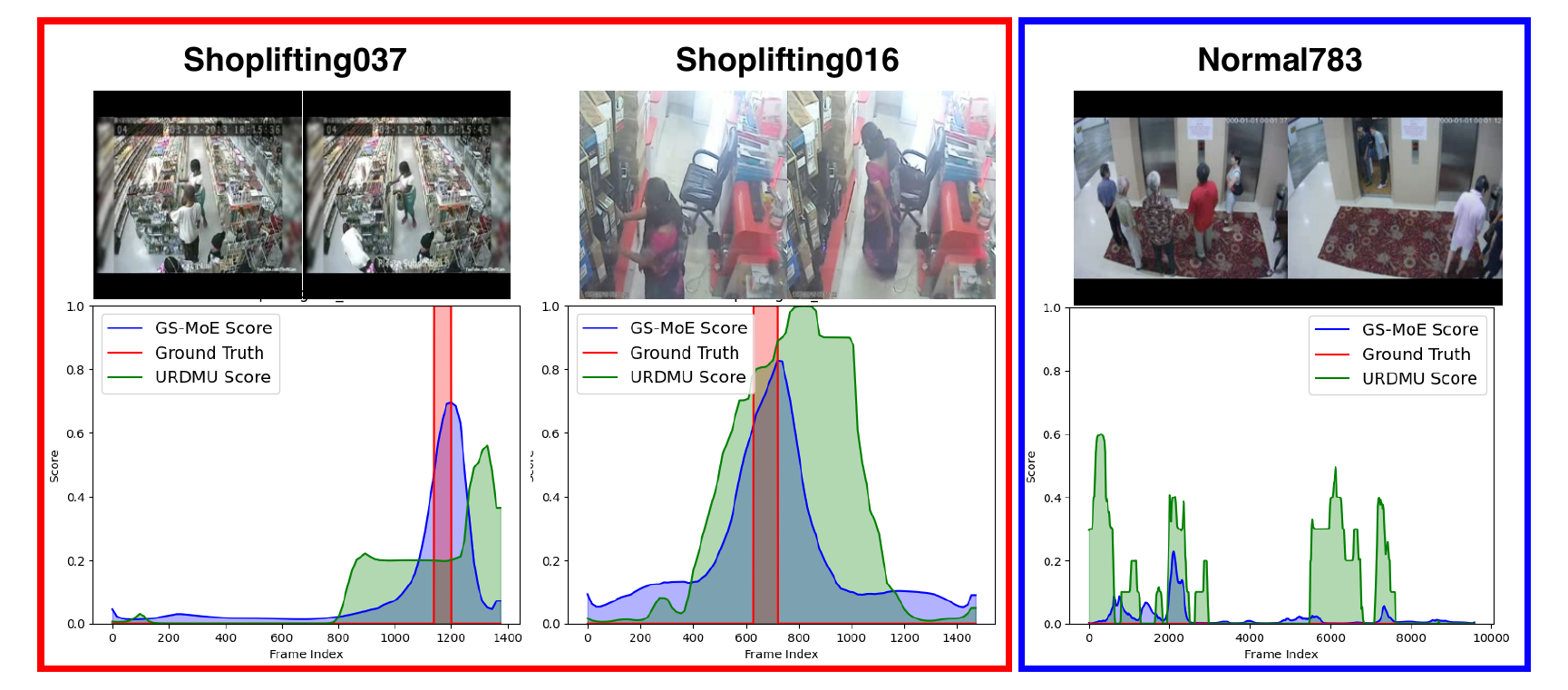}
    \caption{GS-MoE qualitatively outperforms the URDMU model used to produce the task-aware features. Specifically, in the case of subtle anomalies, TGS allows to localize the anomalous event precisely within the ground-truth time window, but the predicted scores have lower values in the peak (true positive) than the ones predicted by the baseline. Similarly, the anomaly scores produced for normal videos (false positives) have lower scores compared to the baseline. }
    \label{fig:qa}
\end{figure}

\section{TGS Ablation Study}
\label{sec:tgs}
As mentioned in Section 3.1, in order to mitigate the presence of spurious peaks, a model trained with TGS has to be warmed up using the standard MIL loss function or the $L_{topk-norm}$ component. 
We conducted experiments with different peaks thresholds to evaluate the sensitivity of our approach to the selection of peaks. As shown in Table~\ref{tab:thresh}, the performance of the model are marginally influenced by the threshold selected within the range of 0.1 and 0.3. For threshold values below 0.1, TGS detects too many peaks, especially in early stages of training, which does not allow the model to converge. On the other hand, a threshold above 0.3 leads to selecting very few peaks, leading the model to estimate low scores for every video due to the fact that the major component of the loss function is given by the $top_k$ normal frames. 

\begin{table}[ht]
    \centering
    \begin{tabular}{c|ccccc}
    \toprule
        & \multicolumn{5}{c}{Threshold} \\
        & 0.1 & 0.15 & 0.2 & 0.25 & 0.3 \\
        \midrule
        AUC & 90.34 & 91.08 & \bf 91.58 & 91.23 & 90.75 \\
        \bottomrule
    \end{tabular}
    \caption{Performance comparison between different peak thresholds on the UCF-Crime dataset.}
    \label{tab:thresh}
\end{table}

\section{Soft MoE}
\label{sec:soft}

In order to provide an overview of the capabilities of the proposed GS-MoE framework, we implement the same training strategy with soft-MoE, a modern MoE architecture  introduced by \citep{puigcerver2024sparsesoftmixturesexperts}. The framework, shown in Figure~\ref{fig:soft_moe}, differs from the Gating model detailed in the main paper by the strategy used to leverage the expert's predictions. In the soft-MoE architecture, the task-aware features are processed by a linear layer followed by a transformer block. A MLP predicts abnormal scores for each anomaly class in the dataset. Subsequently, these abnormal scores are weighted by the abnormal scores predicted by the experts to produce a single abnormal score. 

We conducted experiments with this architecture on the UCF-Crime following the same training strategy detailed in the main paper. The results, reported in Table~\ref{tab:moe_comparison}, show that the Gating model presented in the main paper achieves a 1.44\% higher $AUC$ score compared to soft-MoE. This result is in line with the results reported in Table 3 of the main paper, which highlights the benefits of processing the task-aware features together with the expert's scores.

\begin{table}[ht]
    \centering
    \begin{tabular}{c|cc}
        \toprule
         & Gate & Soft \\
         \midrule
         AUC & \bf 91.58 & 90.14 \\
         \bottomrule
    \end{tabular}
    \caption{Comparison between the Gating MoE and the Soft MoE architectures for GS-MoE. We report the $AUC$ score achieved on the UCF-Crime dataset.}
    \label{tab:moe_comparison}
\end{table}
\paragraph{Computational Costs:}
\label{sec:costs}
GS-MoE increases the computational cost over a SoTA baseline model, while still able to process 10 frames per second. It is important to notice that, in our implementation, the experts process the input in sequence, while a parallel implementation would result in higher fps and near real-time performance.
\begin{table}[ht]
\small
\setlength\tabcolsep{3.5pt} 
    \centering
    \begin{tabular}{c|ccc|c}
    \toprule
         & \bf \small UR-DMU & \bf \small Experts & \small \bf Gate & \small \bf GS-MoE (Our) \\
        \midrule
        \bf GFLOPs & 1.54 & 1.56 & 0.789 & 4.133\\
        \bf Params. (M) & 6.16 & 6.52 & 3.34 & 16.02 \\
        \bf FPS & 110.09 & 35.73 & 212.83 & 9.57\\
        \midrule
        $AUC$ & 86.97 & 89.53 & - & \bf 91.58 \\
        \bottomrule
    \end{tabular}
    \caption{Computational cost analysis for UCF-Crime with 13 experts.}
    \label{tab:computational}
\end{table}
\section{Qualitative results - Failure Cases}
\label{sec:qa}
\begin{figure*}[ht]
    \centering
    \includegraphics[width=\textwidth]{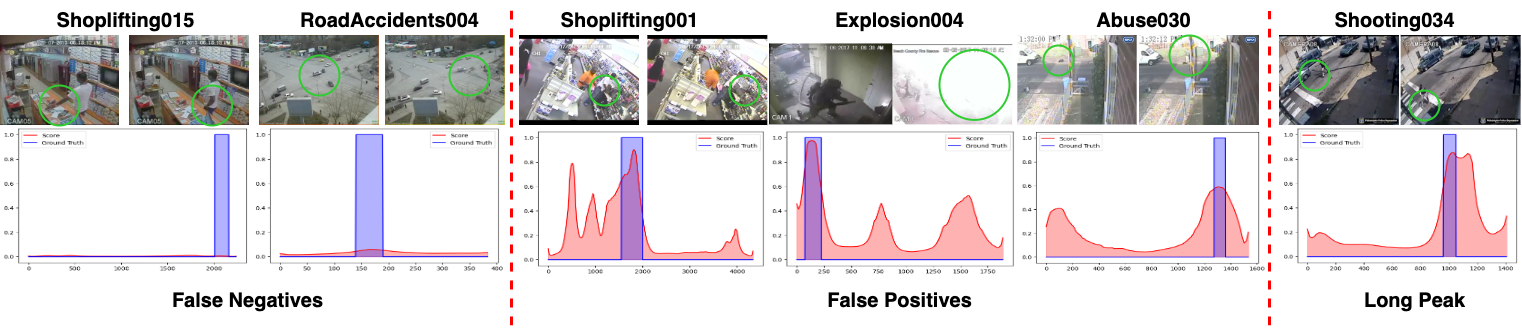}
    \caption{Failure cases examples on the UCF-Crime dataset.}
    \label{fig:failure_qa}
\end{figure*}
In Figure~\ref{fig:failure_qa} we report some examples of videos on which GS-MoE is unable to correctly detect the anomalous portion of the video. We identify three main failure cases: false negative, false positives and long peaks. 

In WSVAD, a false negative is a missed detection of an anomaly in a video. For ``Shoplifting-015" and ``RoadAccidents-004", GS-MoE predicts abnormal scores close to zero for every frame. In the former example, the anomalous action is very subtle and requires a deeper understanding of the context in which the anomaly happens. Additional context cues could be useful in such cases, such as the inclusion of text features via a video-captioning model. On the other hand, in ``RoadAccidents-004" the anomaly happens in a very small pixel-region of the video due to the camera being far away from the scene. 

False positives are instances where GS-MoE predicts (relatively) high abnormal scores for portions of the videos that do not contain anomalous actions. The shape of the false positive peaks in the abnormal scores of ``Explosion-004" and ``Abuse-030" suggests that TGS could be partially responsible for them. On the other hand, in the ``Shoplifting-001" video the frames in the ground-truth anomaly region closely resemble the previously ones and identifying when the anomaly starts is challenging for humans as well. 

In the last example, the anomaly is correctly detected but the peak is extended further outside the anomaly region. In fact in the video of ``Shooting-024", a person can be seen shooting in an empty street and then remaining on the road for a few seconds before entering a vehicle. This seems to be a common issue in videos where the anomaly action has lasting effects on the scene. 

\section{UBnormal Experiments}
\label{sec:ubnormal}
In order to present a more comprehensive overview of the performance of GS-MoE, we experiment on the UBNormal dataset~\cite{Acsintoae_CVPR_2022}. This dataset is composed of synthetic videos generated in 29 different scenes. We experiment on this dataset in order to show the efficiency of our proposed model in data-constrained context. In fact, UBnormal contains 14.02 minutes of abnormal videos and 50.48 minutes of normal videos in the training set. The dataset does not contain anomaly-class labels, therefore we train an expert on normal and abnormal videos of a single scene, obtaining 29 scene-specialized experts. We compare the performance of our GS-MoE with other baseline models in Table~\ref{tab:ubnormal}. We also experiment by clustering the anomalous videos in the training set and assigning an expert to each cluster, as described for the UCF-Crime dataset in Section {\color{red}4.3} of the main paper.

\begin{table}[t]
    \centering
    \begin{tabular}{c|c}
    \toprule
         Model & AUC \\
         \midrule
         Georgescu et al. \citep{georgescu2021background} & 61.3 \\
         Sultani et al. \citep{sultani2018real} & 50.3\\
         Bertasius et al. \citep{bertasius2021space} & 68.5\\
        UR-DMU\citep{AAAI23URDMU}{\color{red}*} & 61.03 \\
         \midrule
        GS-MoE (5 clusters / 5 experts) & 68.50 \\
        GS-MoE (6 clusters / 6 experts) & 67.61\\
        \bf GS-MoE (7 clusters / 7 experts) & \bf 69.28\\
        GS-MoE (8 clusters / 8 experts) & 64.08\\
        GS-MoE (9 clusters / 9 experts) & 67.82\\
        GS-MoE (10 clusters / 10 experts) & 68.87\\
        GS-MoE (11 clusters / 11 experts) & 68.61\\
        GS-MoE (12 clusters / 12 experts) & 68.54\\
        GS-MoE (13 clusters / 13 experts) & 68.82\\
        GS-MoE (14 clusters / 14 experts) & 68.78\\
        GS-MoE (15 clusters / 15 experts) & 68.55\\
         \midrule
        GS-MoE (scene experts) & \bf 65.95 \\
        \bottomrule
    \end{tabular}
    \caption{Performance comparison on the UBnormal dataset. {\color{red}*} indicates our own implementation and was used as task-aware feature extractor.}
    \label{tab:ubnormal}
\end{table}

The training set of UBnormal contains 82 abnormal videos and 186 normal videos in total, but it is important to notice that there are no training abnormal videos for some scenes (scenes 7, 10 and 15), while for others there is only one anomalous video (scenes 1, 2, 5, 13, 17 and 28). This leads to a very unbalanced set of experts for the scene-experts implementation, which strongly hinders the overall performance of GS-MoE. However, GS-MoE achieves 65.95\% on the $AUC$ metric in the scene-experts setting, which is better or on par with baseline methods, highlighting the efficiency of the proposed framework in such a data-constrained setting. 

In the context of this dataset, the cluster-experts do not suffer from the lack of scene-specific anomalies and consistently exhibit much better performance than the scene-expert version. By clustering the training anomalous videos in seven clusters, GS-MoE is able to achieve 69.28\% on the $AUC$ metric, surpassing most baseline methods albeit falling short of the SoTA mark.

\end{document}